\documentclass{article} 
\usepackage{iclr2026_conference,times}

\usepackage{amsmath,amsfonts,bm}

\def\eqref#1{equation~\ref{#1}}

\def\1{\bm{1}}

\DeclareMathAlphabet{\mathsfit}{\encodingdefault}{\sfdefault}{m}{sl}
\SetMathAlphabet{\mathsfit}{bold}{\encodingdefault}{\sfdefault}{bx}{n}

\usepackage{hyperref}
\usepackage{url}
\usepackage{amsmath, amssymb}
\usepackage{enumitem}

\usepackage{graphicx}
\usepackage{algorithm}
\usepackage{algpseudocode}
\usepackage{booktabs}
\usepackage{amsmath}
\usepackage{amsfonts}
\usepackage[us,12hr]{datetime}
\usepackage[table]{xcolor}
\usepackage{listings}
\usepackage{listingsutf8}
\usepackage{wrapfig}
\usepackage{multicol}
\usepackage{caption}
\usepackage{mdframed}
\usepackage[utf8]{inputenc}
\usepackage[most]{tcolorbox}

\lstdefinelanguage{lean} {%
mathescape=true,
texcl=false,
morekeywords=[1]{
import, prelude, protected, private, noncomputable, definition, meta, renaming,
hiding, exposing, parameter, parameters, begin, conjecture, constant, constants,
hypothesis, lemma, corollary, variable, variables, premise, premises, theory,
print, theorem, proposition, example, abstract,
open, as, export, override, axiom, axioms, inductive, with, without,
structure, record, universe, universes,
alias, help, precedence, reserve, declare_trace, add_key_equivalence,
match, infix, infixl, infixr, notation, postfix, prefix, instance,
eval, vm_eval, check, coercion, end, this, suppose,
using, using_well_founded, namespace, section, fields,
attribute, local, set_option, extends, include, omit, classes, class,
instances, coercions, attributes, raw, replacing,
calc, have, show, suffices, by, in, at, let, forall, Pi, fun,
exists, if, dif, then, else, assume, take, obtain, from, aliases, register_simp_ext,
mutual, def, run_command
},
morekeywords=[2]{Type, Prop, Type*, Type₀, Type₁, Type₂, Type₃},
literate=
{α}{{\ensuremath{\mathrm{\alpha}}}}1
{β}{{\ensuremath{\mathrm{\beta}}}}1
{γ}{{\ensuremath{\mathrm{\gamma}}}}1
{δ}{{\ensuremath{\mathrm{\delta}}}}1
{ε}{{\ensuremath{\mathrm{\varepsilon}}}}1
{ζ}{{\ensuremath{\mathrm{\zeta}}}}1
{η}{{\ensuremath{\mathrm{\eta}}}}1
{θ}{{\ensuremath{\mathrm{\theta}}}}1
{ι}{{\ensuremath{\mathrm{\iota}}}}1
{κ}{{\ensuremath{\mathrm{\kappa}}}}1
{μ}{{\ensuremath{\mathrm{\mu}}}}1
{ν}{{\ensuremath{\mathrm{\nu}}}}1
{ξ}{{\ensuremath{\mathrm{\xi}}}}1
{π}{{\ensuremath{\mathrm{\mathnormal{\pi}}}}}1
{ρ}{{\ensuremath{\mathrm{\rho}}}}1
{σ}{{\ensuremath{\mathrm{\sigma}}}}1
{τ}{{\ensuremath{\mathrm{\tau}}}}1
{φ}{{\ensuremath{\mathrm{\varphi}}}}1
{χ}{{\ensuremath{\mathrm{\chi}}}}1
{ψ}{{\ensuremath{\mathrm{\psi}}}}1
{ω}{{\ensuremath{\mathrm{\omega}}}}1
{Γ}{{\ensuremath{\mathrm{\Gamma}}}}1
{Δ}{{\ensuremath{\mathrm{\Delta}}}}1
{Θ}{{\ensuremath{\mathrm{\Theta}}}}1
{Λ}{{\ensuremath{\mathrm{\Lambda}}}}1
{Σ}{{\ensuremath{\mathrm{\Sigma}}}}1
{Φ}{{\ensuremath{\mathrm{\Phi}}}}1
{Ξ}{{\ensuremath{\mathrm{\Xi}}}}1
{Ψ}{{\ensuremath{\mathrm{\Psi}}}}1
{Ω}{{\ensuremath{\mathrm{\Omega}}}}1
{ℵ}{{\ensuremath{\aleph}}}1
{≤}{{\ensuremath{\leq}}}1
{≥}{{\ensuremath{\geq}}}1
{≠}{{\ensuremath{\neq}}}1
{≈}{{\ensuremath{\approx}}}1
{≡}{{\ensuremath{\equiv}}}1
{≃}{{\ensuremath{\simeq}}}1
{≤}{{\ensuremath{\leq}}}1
{≥}{{\ensuremath{\geq}}}1
{∂}{{\ensuremath{\partial}}}1
{∆}{{\ensuremath{\triangle}}}1 
{∫}{{\ensuremath{\int}}}1
{∑}{{\ensuremath{\mathrm{\Sigma}}}}1
{Π}{{\ensuremath{\mathrm{\Pi}}}}1
{⋂}{{\ensuremath{\cap}}}1
{⊥}{{\ensuremath{\perp}}}1
{∞}{{\ensuremath{\infty}}}1
{∂}{{\ensuremath{\partial}}}1
{∓}{{\ensuremath{\mp}}}1
{±}{{\ensuremath{\pm}}}1
{×}{{\ensuremath{\times}}}1
{⊕}{{\ensuremath{\oplus}}}1
{⊗}{{\ensuremath{\otimes}}}1
{⊞}{{\ensuremath{\boxplus}}}1
{∇}{{\ensuremath{\nabla}}}1
{√}{{\ensuremath{\sqrt}}}1
{⬝}{{\ensuremath{\cdot}}}1
{•}{{\ensuremath{\cdot}}}1
{∘}{{\ensuremath{\circ}}}1
{⁻}{{\ensuremath{^{-}}}}1
{▸}{{\ensuremath{\blacktriangleright}}}1
{∧}{{\ensuremath{\wedge}}}1
{∨}{{\ensuremath{\vee}}}1
{¬}{{\ensuremath{\neg}}}1
{⊢}{{\ensuremath{\vdash}}}1
{⟨}{{\ensuremath{\langle}}}1
{⟩}{{\ensuremath{\rangle}}}1
{↦}{{\ensuremath{\mapsto}}}1
{→}{{\ensuremath{\rightarrow}}}1
{↔}{{\ensuremath{\leftrightarrow}}}1
{⇒}{{\ensuremath{\Rightarrow}}}1
{⟹}{{\ensuremath{\Longrightarrow}}}1
{⇐}{{\ensuremath{\Leftarrow}}}1
{⟸}{{\ensuremath{\Longleftarrow}}}1
{∩}{{\ensuremath{\cap}}}1
{∪}{{\ensuremath{\cup}}}1
{⊂}{{\ensuremath{\subseteq}}}1
{⊆}{{\ensuremath{\subseteq}}}1
{⊄}{{\ensuremath{\nsubseteq}}}1
{⊈}{{\ensuremath{\nsubseteq}}}1
{⊃}{{\ensuremath{\supseteq}}}1
{⊇}{{\ensuremath{\supseteq}}}1
{⊅}{{\ensuremath{\nsupseteq}}}1
{⊉}{{\ensuremath{\nsupseteq}}}1
{∈}{{\ensuremath{\in}}}1
{∉}{{\ensuremath{\notin}}}1
{∋}{{\ensuremath{\ni}}}1
{∌}{{\ensuremath{\notni}}}1
{∅}{{\ensuremath{\emptyset}}}1
{∖}{{\ensuremath{\setminus}}}1
{†}{{\ensuremath{\dag}}}1
{ℕ}{{\ensuremath{\mathbb{N}}}}1
{ℤ}{{\ensuremath{\mathbb{Z}}}}1
{ℝ}{{\ensuremath{\mathbb{R}}}}1
{ℚ}{{\ensuremath{\mathbb{Q}}}}1
{ℂ}{{\ensuremath{\mathbb{C}}}}1
{⌞}{{\ensuremath{\llcorner}}}1
{⌟}{{\ensuremath{\lrcorner}}}1
{⦃}{{\ensuremath{\{\!|}}}1
{⦄}{{\ensuremath{|\!\}}}}1
{₁}{{\ensuremath{_1}}}1
{₂}{{\ensuremath{_2}}}1
{₃}{{\ensuremath{_3}}}1
{₄}{{\ensuremath{_4}}}1
{₅}{{\ensuremath{_5}}}1
{₆}{{\ensuremath{_6}}}1
{₇}{{\ensuremath{_7}}}1
{₈}{{\ensuremath{_8}}}1
{₉}{{\ensuremath{_9}}}1
{₀}{{\ensuremath{_0}}}1
{¹}{{\ensuremath{^1}}}1
{⌊}{{\ensuremath{\lfloor}}}1
{⌋}{{\ensuremath{\rfloor}}}1
{ₙ}{{\ensuremath{_n}}}1
{ₘ}{{\ensuremath{_m}}}1
{↑}{{\ensuremath{\uparrow}}}1
{↓}{{\ensuremath{\downarrow}}}1
{▸}{{\ensuremath{\triangleright}}}1
{Σ}{{\color{symbolcolor}\ensuremath{\Sigma}}}1
{Π}{{\color{symbolcolor}\ensuremath{\Pi}}}1
{∀}{{\color{symbolcolor}\ensuremath{\forall}}}1
{∃}{{\color{symbolcolor}\ensuremath{\exists}}}1
{λ}{{\color{symbolcolor}\ensuremath{\mathrm{\lambda}}}}1
{:=}{{\color{symbolcolor}:=}}1
{=}{{\color{symbolcolor}=}}1
{<}{{\color{symbolcolor}<}}1
{+}{{\color{symbolcolor}+}}1
{*}{{\color{symbolcolor}*}}1,
morecomment=[s][\color{commentcolor}]{/-}{-/},
morecomment=[l][\itshape \color{commentcolor}]{--},
showstringspaces=false,
keepspaces=true,
morestring=[b]",
morestring=[d],
tabsize=3,
extendedchars=true,
sensitive=true,
breaklines=true,
basicstyle=\ttfamily,
captionpos=b,
columns=[l]fullflexible,
identifierstyle={\ttfamily\color{black}},
keywordstyle=[1]{\ttfamily\color{keywordcolor}},
keywordstyle=[2]{\ttfamily\color{sortcolor}},
stringstyle=\ttfamily,
}

\definecolor{keywordcolor}{rgb}{0.7, 0.1, 0.1}   
\definecolor{commentcolor}{rgb}{0.4, 0.4, 0.4}   
\definecolor{symbolcolor}{rgb}{0.0, 0.1, 0.6}    
\definecolor{sortcolor}{rgb}{0.1, 0.5, 0.1}      
\definecolor{errorcolor}{rgb}{1, 0, 0}           
\definecolor{stringcolor}{rgb}{0.5, 0.3, 0.2}    

\definecolor{promptbg}{gray}{0.95}
\definecolor{promptborder}{gray}{0.7}

\lstdefinestyle{leanstyle}{
    language=lean,
    basicstyle=\ttfamily\footnotesize,
    keywordstyle=\color{blue}\bfseries,
    commentstyle=\color{gray},
    stringstyle=\color{red},
    numberstyle=\tiny\color{gray},
    numbers=left,
    numbersep=5pt,
    frame=single,
    frameround=tttt,
    framesep=10pt,
    rulecolor=\color{black},
    backgroundcolor=\color{white},
    breaklines=true,
    breakatwhitespace=false,
    tabsize=2,
    captionpos=b,
    aboveskip=10pt,
    belowskip=10pt,
    xleftmargin=10pt,
    xrightmargin=10pt
}
\lstdefinestyle{promptstyle}{
    basicstyle=\ttfamily\footnotesize,
    backgroundcolor=\color{promptbg},
    frame=single,
    frameround=tttt,
    framerule=0.5pt,
    rulecolor=\color{promptborder},
    breaklines=true,
    breakatwhitespace=false,
    breakindent=0pt,
    showspaces=false,
    showstringspaces=false,
    showtabs=false,
    tabsize=2,
    captionpos=b,
    numbers=none,
    lineskip=1pt
}
\lstdefinestyle{lean}{
  basicstyle=\ttfamily\small,
  keywordstyle=\color{blue}\bfseries,
  commentstyle=\color{gray},
  stringstyle=\color{red},
  columns=fullflexible,
  frame=single,
  keepspaces=true,
  escapeinside={(*}{*)},
  morekeywords={theorem, fun, by, sorry, Even}
}

\lstdefinestyle{leanex}{
  basicstyle=\ttfamily\small,
  keywordstyle=\color{blue}\bfseries,
  commentstyle=\color{gray},
  stringstyle=\color{red},
  columns=fullflexible,
  keepspaces=true,
  escapeinside={(*}{*)},
  morekeywords={theorem, fun, by, sorry, Even}
}

\lstdefinestyle{problem}{
  basicstyle=\rmfamily\small,
  frame=single,
  columns=fullflexible,
  escapeinside={(*}{*)},
  breaklines=true
}

\newenvironment{prompt}[1][]{%
  \begin{tcolorbox}[colback=black!5, colframe=black!50, 
                    sharp corners, boxrule=0.5pt,
                    fontupper=\ttfamily,
                    breakable,
                    enhanced,
                    title=#1]%
}{%
  \end{tcolorbox}%
}

\definecolor{promptbg}{RGB}{245, 245, 250}
\definecolor{promptborder}{RGB}{100, 100, 150}
\definecolor{prompttext}{RGB}{50, 50, 100}

\usepackage{wrapfig}

\title{\textsc{IndiMathBench}: Autoformalizing Mathematical Reasoning Problems with a Human Touch}

\author{
  Param Biyani \quad
  Shashank Kirtania \quad
  Yasharth Bajpai \quad
  Sumit Gulwani \quad
  Ashish Tiwari \\
  \parbox{\linewidth}{\centering\normalsize{Microsoft}}\\
  \parbox{\linewidth}{\centering\texttt{parambiyani8@gmail.com, t-shkirtania@microsoft.com}}
}

\newcommand{\benchmark}{\textsc{IndiMathBench}}

\newcommand{\param}[1]{{\textcolor{blue}{#1}}}

\iclrfinalcopy 

\begin{document}

\maketitle
\begin{abstract}

Reliable autoformalization remains challenging even in the era of large language models (LLMs). 
%
%
%
The scarcity of high-quality training data is a major bottleneck. Expert annotation requires substantial time and deep expertise in both mathematics and theorem proving.
%
%
%
We introduce \benchmark, a human-verified benchmark designed to evaluate mathematical theorem proving, curated using an AI-powered human-assisted pipeline for formalizing natural language problems in Lean.
\benchmark~is composed of 312 formal Lean 4 theorems paired with their corresponding informal problem statements, sourced from Indian Mathematics Olympiads.
Through category-based retrieval, iterative compiler feedback, and multi-model ensembles, our pipeline generates candidate formalizations that experts efficiently validate via an interactive dashboard with automated quality summaries.
%
%
Evaluation across multiple frontier models demonstrates that autoformalization remains challenging, with substantial gaps between syntactic validity and semantic correctness, while theorem proving success rates remain low even with iterative refinement, demonstrating that \benchmark~presents a challenging testbed for mathematical reasoning.
\benchmark~is available at \href{https://github.com/prmbiy/IndiMathBench}{https://github.com/prmbiy/IndiMathBench}.

\end{abstract}

\newcommand\ignore[1]{{}}
\section{Introduction}

The formalization of mathematics, expressing informal reasoning in precise, machine-verifiable logic, has been a long-standing goal in computer science and mathematics.
A related task, autoformalization, seeks to automatically translate informal mathematical statements and proofs into formal representations \citep{autoformalization_llm, autoformalization_codex_1, autoformalization_codex_2}.
Despite advances in large language models and theorem-proving frameworks such as Lean \citep{lean}, progress remains limited by the scarcity of paired informal–formal data.

\setlength{\intextsep}{0.5em}
\begin{wrapfigure}{r}{0.5\textwidth}
\vspace{-0.1in}
\begin{mdframed}[linewidth=0.5pt,innerleftmargin=6pt,innerrightmargin=6pt,innertopmargin=6pt,innerbottommargin=6pt]
\small
Let n be a natural number. Prove that:
\[\left\lfloor \frac{n}{1} \right\rfloor + \left\lfloor \frac{n}{2} \right\rfloor + \left\lfloor \frac{n}{3} \right\rfloor + \cdots + \left\lfloor \frac{n}{n} \right\rfloor + \left\lfloor \sqrt{n} \right\rfloor\]
is even.
\end{mdframed}
\vspace{0.05em}
\begin{mdframed}[linewidth=0.5pt,innerleftmargin=6pt,innerrightmargin=6pt,innertopmargin=0pt,innerbottommargin=0pt]
\begin{lstlisting}[style=leanex]
theorem inmo_2014_2 (n : (*$\mathbb{N}$*)) :
  Even ((Finset.sum (Finset.range n)
    fun i => (*$\lfloor$*)(n : (*$\mathbb{R}$*)) / ((i + 1) : 
    (*$\mathbb{R}$*))(*$\rfloor$*)) + (*$\lfloor$*)Real.sqrt n(*$\rfloor$*)) := by
  sorry 
\end{lstlisting}
\end{mdframed}
\vspace{-0.1in}
\caption{A sample problem from INMO 2014 with its formalization in Lean 4}
\label{fig:inmo_2014_4}
\vspace{-0.1in}
\end{wrapfigure}


Existing benchmarks for formal theorem proving are few and narrow in scope.
The largest Lean 4 benchmarks with Olympiad-level problems, \textsc{miniF2F}~\citep{zheng2022minif2f}, drawn from AIME, AMC, and IMO exams, and PutnamBench~\citep{tsoukalas2024putnambench}, from the Putnam Competition, cover only a small fraction of the available competition mathematics, totaling roughly a thousand problems. This limited scale restricts comprehensive evaluation of model generalization and reasoning capabilities.

Moreover, these popular sources increasingly appear in large-scale pretraining corpora \citep{jiang2024investigatingdatacontaminationpretraining}, creating contamination that obscures genuine reasoning ability.
Producing new benchmarks is further constrained by the high manual effort required: experts must formalize, annotate, and verify each problem in Lean, a process that is both time-consuming and resource-intensive~\citep{yu2025formalmathbenchmarkingformalmathematical}. As a result, progress in ATP evaluation depends on developing scalable, high-quality, human-verified formal benchmarks.
Also, despite LLMs’ improved syntax fidelity, their semantic alignment with mathematical intent remains poor. We hypothesize that human-AI collaborative formalization can bridge this gap.



To address these challenges, we introduce \benchmark~, a human-verified benchmark for automated theorem proving built from Indian Mathematical Olympiad problems, built using a human-AI pipeline.
Our benchmark contains 312 problems spanning diverse mathematical domains, geometry, algebra, number theory, and combinatorics, each paired with human-verified Lean 4 formalizations.
A sample problem from \benchmark~is shown in Figure~\ref{fig:inmo_2014_4}.
We conducted systematic human verification of all formalizations, ensuring high-quality ground truth for reliable evaluation.

Our key contributions are:
\begin{itemize}\itemsep=0em
\item A Lean 4 benchmark for formal theorem proving, created using LLM-assisted formalization and human verification.
\item A systematic formalization pipeline employing category-based retrieval, a self-debug loop, with an ensemble of LLMs, and cross comparison of formalizations and a controlled study on its efficacy gains.
\item A VS Code Extension for improved human-AI collaboration for Lean annotations.
\item Evaluation of frontier general purpose and fine-tuned models on their autoformalization and proving capabilities.
\end{itemize}

\section{Related Work}

\textbf{Formal Theorem Proving Benchmarks.} 
The evaluation of automated theorem proving systems relies critically on curated formal benchmarks, yet existing resources remain limited in scale, diversity, and representativeness. The \textsc{miniF2F} benchmark \citep{zheng2022minif2f}, based on AMC, AIME, and IMO competitions, and PutnamBench \citep{tsoukalas2024putnambench}, derived from the Putnam Competition, have become standard datasets for assessing Olympiad competition-level mathematical reasoning in Lean 4.
More recent benchmarks such as ProofNet \citep{azerbayev2023proofnet} and FormalMath \citep{yu2025formalmathbenchmarkingformalmathematical} focus on undergraduate or textbook-style mathematics, providing complementary but less challenging problem domains.
While these resources have advanced systematic evaluation, they remain narrow in both mathematical coverage and cultural scope, largely reflecting Western curricula and competition traditions, where the focus is relatively less on geometry and combinatorics style problems, and more towards analysis, algebra and number theory.
Recent efforts like FrontierMath \citep{frontiermath2024} extend evaluation toward research-level problems but reveal extremely low model success rates, underscoring the substantial gap between current benchmarks and the diversity of human mathematical reasoning, but doesn't support Lean.

In particular, Olympiad-style geometry and combinatorics remain strikingly underrepresented across all existing Lean 4 benchmarks. Even the most widely used Olympiad datasets contain only a small number of high-difficulty geometry problems, despite these being among the hardest domains for both ATPs and LLM-guided provers. Recent efforts such as LeanGeo \citep{song2025leangeoformalizingcompetitionalgeometry} and CombiBench \citep{liu2025combibenchbenchmarkingllmcapability} were motivated by this same gap, indicating that it is a well-recognized limitation in the formalization ecosystem for geometry and combinatorics respectively. This imbalance limits the ability of current benchmarks to stress-test systems on categories that are disproportionately challenging.

\textbf{Autoformalization.} 
Autoformalization aims to translate informal mathematical language into machine-verifiable formal logic (like Lean). Recent advances in large language models have substantially improved the ability to generate structured formal statements directly from natural language \citep{autoformalization_llm}.
Despite these advances, fully automated approaches continue to struggle with semantic consistency, complex mathematical reasoning, and the domain gap between natural language and formal specifications \citep{zheng2022minif2f, welleck2022naturalprover, kirtania2025steeringllmsformaltheorem}. Current methods struggle to preserve precise semantic details when translating natural-language problems into Lean, particularly in advanced domains, limiting the availability of reliable formal reasoning benchmarks. 

Another challenge is the evaluation of these autoformalization systems. Syntactic approaches like BLEU score \citep{bleu} do not consider the semantic distance between generations. Recent methods like Generalized Tree Edit Distance (GTED) \citep{liu2025gted} compute structural similarity through operator trees, while Bidirectional Equivalence (BEq) \citep{liu2025rethinking} provides neural-symbolic equivalence checking. Both report high correlation with human evaluations. Other methods include combining certainty and similarity scores through contrastive learning \citep{lu2024formalalign}. However, scalable evaluation remains challenging, with manual expert verification being costly but necessary for establishing reliable ground truth.

\textbf{Human-AI Collaboration in Formalization.}
As human verification remains a necessity, other Lean 4 benchmarks like \textsc{miniF2F}, PutnamBench, and ProofNet use completely manual annotation for robust formalizations; the process is costly and non-scalable.
However, the formalization of research-level mathematics increasingly relies on human-AI collaboration that combines automated translation capabilities with human expertise in both competition mathematics and theorem-proving languages \citep{fimo, yu2025formalmathbenchmarkingformalmathematical}.
Process-supervised approaches use formal system feedback to improve translation quality while reducing annotation requirements. These hybrid methodologies outperform pure automation by combining human expertise for conceptual insights with AI capabilities for pattern recognition and verification. However, no prior work has systematically evaluated these hybrid pipelines through a controlled study to measure their actual contribution to formalization quality and annotation efficiency. This gap limits understanding of when and how human–AI collaboration provides measurable value, and prevents principled design of future human-AI pipelines on complex reasoning tasks.

\begin{figure}[t]
    \centering
    \includegraphics[width=\textwidth]{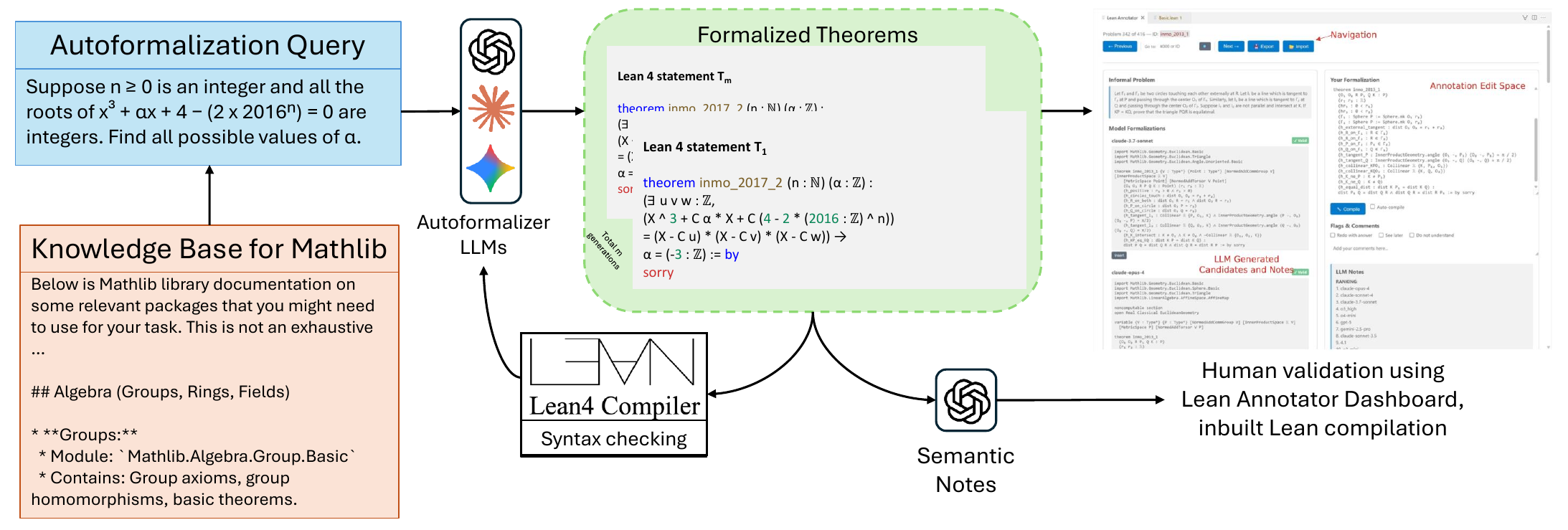}
    \caption{{\small{Overview of approach for creating IndicMath dataset: Human is assisted by a multiple LLM annotators. Each LLM generation is conditioned on the natural language and documentation, and goes through a validation check by Lean. Errors are provided as feedback in subsequent iterations. The final generations from all LLMs are summarized by an LLM in a dashboard to help optimize annotation efficiency.}}}
    \label{fig:pipeline_overview}
\end{figure}

\section{\textsc{\benchmark~}}
\begin{wraptable}{r}{0.4\textwidth}
\centering
\small
\begin{tabular}{l r}
\textbf{Category} & \textbf{Count} \\
\hline
Geometry & 98 \\
Algebra & 92 \\
Set Theory \& Combinatorics & 45 \\
Number Theory & 77 \\
\hline
\rowcolor{gray!20}\textbf{Total} & \textbf{312} \\
\hline
\end{tabular}
\caption{{\small{Problem distribution by topic domain in \benchmark.}}}\label{table:categories}
\label{table1:problem_domains}
\end{wraptable}

\benchmark~contains 312 formalized problem statements from Indian Olympiad problems. 
Each problem consists of an informal problem and an informal proof (in English) of an Olympiad math problem, a Lean 4 theorem corresponding to that problem, and any numerical solution where applicable.
The informal problems are compiled directly from the Regional Mathematical Olympiad (RMO) and Indian National Mathematical Olympiad (INMO) examinations in India.
Table~\ref{table1:problem_domains} gives the number of
problems covering various mathematical domains namely geometry, algebra, number theory, and combinatorics. Figure \ref{fig:pipeline_overview} describes the overall human-AI collaborative procedure for formalization.

\textbf{Diversity and Breadth.} Compared to existing benchmarks like \textsc{miniF2F} (Zheng et al., 2021), which include a wide selection of high-school and early undergraduate mathematics problems (primarily from AMC, AIME, and IMO), the \benchmark~focuses on problems from the Indian Mathematical Olympiad system. These problems are drawn from national and regional competitions in India, and are restricted to the high-school curriculum, covering algebra, number theory, geometry, and combinatorics. Unlike \textsc{miniF2F}, which incorporates problems involving topics such as inequalities, calculus, or matrix algebra, INMO/RMO problems exclude calculus and typically avoid higher-level abstractions.

Despite this narrower domain scope, INMO problems in particular demonstrate high internal diversity and depth. Many problems involve multi-step reasoning, uncommon constructions, and non-standard techniques. For example, geometric problems often require diagrammatic insight combined with multiple auxiliary constructions, while number-theoretic questions tend to involve clever use of parity, bounding, or invariant arguments.

\textbf{Problem Domains.} 
\benchmark~problems are traditionally sourced from a fixed set of topics: algebra, euclidean geometry, elementary number theory, and combinatorics. Calculus, set theory, and linear algebra are not part of the official syllabus. This restriction makes the benchmark more uniform in scope, but allows for deeper exploration of problem-solving within each domain. For example, geometry problems frequently involve classical triangle centers (e.g., orthocenter, centroid) or cyclic quadrilaterals, which require nontrivial formalization in Lean. Geometry problems which are often under-represented by other datasets due to lack of representation in exams and the lack of support in the Lean + Mathlib ecosystem \citep{song2025leangeoformalizingcompetitionalgeometry}. \benchmark~complements existing benchmarks by drawing from geometry and combinatorics  rich problem sources and formalizing them without using euclidean coordinate representation. 

\textbf{Formalization Effort.}
The formalization was done over the course of a month by two annotators with prior experience in using Lean for formal proof writing. Every annotation was double checked by the other human annotator. In case of difficult problems, or disagreements over the style or correctness of the annotations, the annotators discussed the theorem and mutually came to a final answer. A common code style was followed to make the entire problem be represented within a single theorem statement as possible, and use similar constructions for similar concepts across the benchmark.
AI was used extensively throughout the process and we discuss it in depth in Section \ref{sec:methodology}. 

Some problems (36\%) in our set require solving for a value (rather than proving a statement). 
For these cases, we rephrase the problems to prove for the particular solution similar to prior works like \textsc{miniF2F}, i.e. re-frame ``solve Question Q for Solution X'' as a ``prove Question Q iff Solution X''. An example for such problem can be seen in Appendix \ref{appendix:setup}. Each problem is expressed as a Lean theorem with the proof term replaced by \texttt{sorry}. In Lean, \texttt{sorry} is a placeholder that allows incomplete proofs to compile.


\section{Auto-Formalization Approach}
\label{sec:methodology}

We present a scalable approach for using general-purpose LLMs to accelerate human annotation in the formalization of mathematical problems. This approach, which we used to create \benchmark, is general-purpose and built for reusability.
Algorithm~\ref{alg:main} presents the complete pipeline from natural language problem statements to a final human annotation dashboard.


\subsection{Automated Formalization Generation} 
\label{sec:pipeline}


In our initial evaluations of  for autoformalization, the main deficiency observed was the poor quality of formulas written in custom formal languages. Models often hallucinate non-existent imports, mix syntax from other theorem provers (including Lean 3), and produce ill-formed code. Providing access to library snippets and documentation as feedback notably improves formula correctness, especially for geometry problems where mathlib lacks many competition-level constructs. This inability to consistently follow formal syntax motivates a structured process with documentation access and iterative feedback.

Algorithm~\ref{alg:main} presents the pseudocode for our overall pipeline from problem to human-evaluable dashboard, comprising three key steps:
(a) category-based retrieval, 
(b) iterative refinement with compiler feedback, and 
(c) multi-model ensemble generation with comparative analysis.
All prompts can be found in Appendix \ref{prompt:kb} and \ref{prompt:refinement_prompts}.

\ignore{
\begin{figure*}[t]
    \centering
    \includegraphics[width=\textwidth]{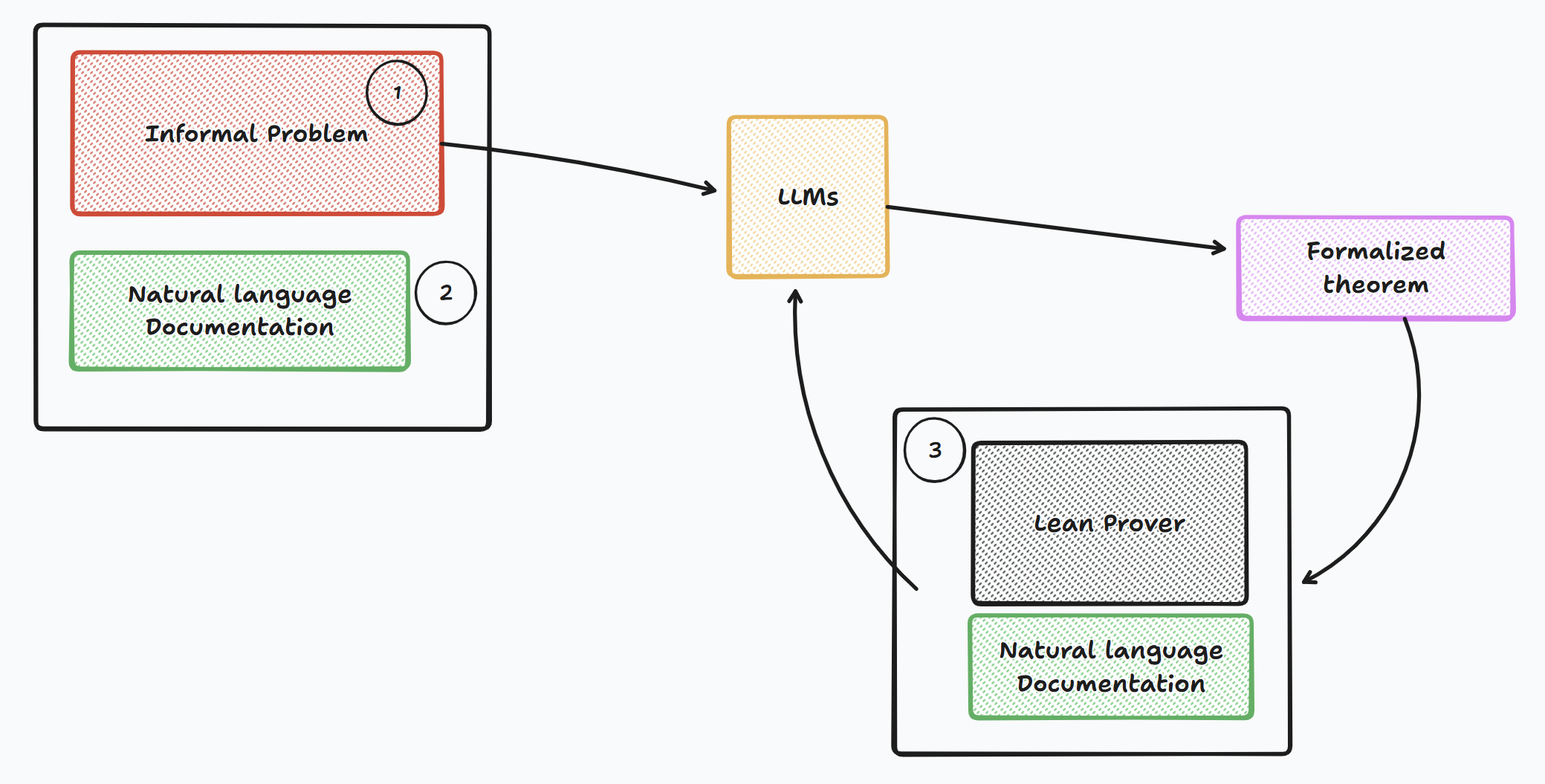}
    \caption{Three setups for autoformalization using large language models (LLMs). This figure illustrates three different setups for leveraging large language models (LLMs) in the task of autoformalization. In the first setup, the LLM is prompted solely with an informal mathematical problem statement. This baseline approach tests the model's ability to directly translate natural mathematical language into a formalized theorem. The second setup augments the informal problem with additional natural language documentation. This supplementary context can include definitions, related lemmas, or domain-specific background, thereby providing the LLM with richer semantic grounding to improve the accuracy and completeness of the formalization. In the third setup, the process is further enhanced by incorporating a formal proof assistant, such as Lean, into the loop. Here, the LLM’s output is checked and iteratively refined based on feedback from the Lean prover, possibly alongside the accompanying natural language documentation. This interaction creates a feedback loop where errors can be detected and corrected, leading to more robust and verifiable formalizations.}
    \label{fig:autoformalization-setups}
\end{figure*}
\endignore}



\ignore{We propose a scalable methodology for leveraging general-purpose LLMs to maximize human annotation efficiency in the formalization of mathematical problems. \ref{fig:pipeline_overview} Semantic correctness requires deep understanding of Olympiad-style mathematics, while syntactic correctness demands precision in theorem-proving languages—together making the process time-intensive and bottlenecked by scarce expertise. To maximize automation, we design a three-stage pipeline. First, we construct a task-specific knowledge base from problem statements and relevant retrieved documentation, where the model, equipped with tools in an agentic setup, explores the Mathlib library and its documentation. Second, generate candidate formalizations by applying Lean-based validation and iterative feedback to ensure syntactic soundness. Third, we employ an ensemble of large language models combined with summarization to consolidate diverse outputs into higher-quality formalizations.}


\ignore{
\paragraph{need to frame this better:}
basically using llms for autoformalization pretty terrible since half-nonsense, hallucinated imports and theorems, and sorry filled compiled failing slop.
So we fix the hallucination by giving it a context map of all the important and relevant stuff, which was also completely automated btw. Less struggle with defining things like circumcenter, incenter etc now. Iterative loops helps it build a proper compiling lean theorem not just slop.
This helps the human annotator have a rough conversion of the problem before they can add the missed out conditions, fix semantics quickly. for results we can show difference in zero shot validity to iterative documentation aided generation. Also an example with same problem same model giving two contrasting results
\endignore}


\vspace{-0.3cm}
\paragraph{Step 1: Category-based Retrieval.}
LLMs often fail to import correct modules or use proper notation in Lean. To mitigate this, we augment prompts with automatically curated mathlib documentation. As described in Procedure~\texttt{PreProcess} (Algorithm~\ref{alg:main}), each problem in the dataset $\mathcal{P}$ is labeled, using an LLM with human verification, into each one of the four categories listed in Table~\ref{table:categories}. We sample 25\% of problems per category and use a Claude Sonnet 4–based agent with bash access to the mathlib repository. The agent explores relevant library code and extracts key definitions and formulas, forming a {\em static context} for each category. This is a one-time, category-specific process: the agent constructs the static context once per domain, which is then reused for all problems in that category to provide consistent access to relevant Mathlib definitions and enhance the quality and clarity of subsequent formalization prompts.
This context anchors the model with domain knowledge within each category, preventing hallucinations and syntax errors.
\vspace{-0.3cm}
\paragraph{Step 2: Iterative Refinement with Error Feedback.}

Figure~\ref{fig:pipeline_overview} illustrates our main formalization loop (Procedure~$\mathtt{Formalize}$). For each problem $p$, the LLM generates a formal theorem $f$ conditioned on the corresponding static context. For “solve”-type problems, we also provide the solution and ask the LLM to generate a verifying theorem. The output is then compiled in Lean 4 via $\mathtt{ValidateInLean}$; any parse or type errors are extracted and fed back to the model for correction. This refinement continues for up to six iterations, yielding syntactically valid Lean formalizations for 95.3\% of problems.

\begin{algorithm}[t]
 \small
    \caption{LLM-Based Autoformalization of Mathematical Problems}\label{alg:main}
    \vspace{-1.5em}
    \begin{multicols}{2}
\begin{algorithmic}[1]
    \Procedure{Formalize}{$p, \mathtt{Model}$} 
    \State \Comment{$p$: problem description in NL}
    %
        %
    \State $ctxt \gets p.Cat.Ctxt$
        \State $f \gets \Call{Model}{p, ctxt}$
        
        \For{$i = 1$ to $6$}
            \State $errors \gets \Call{ValidateInLean}{f}$
            \If{$errors = \emptyset$}
                \State \textbf{break}
            \EndIf
            \State $feedback \gets \Call{ParseErrors}{errors}$
            \State $f \gets \Call{Model}{p, ctxt, f, feedback}$
        \EndFor
    %
    \State \Return $f$
\EndProcedure
\Procedure{Main}{$\mathcal{P},\mathtt{Categories}$} 
    \State $\Call{PreProcess}{\mathcal{P},\mathtt{Categories}}$
    \State $\Call{PostDisplay}{p}$ for $p\in\mathcal{P}$
\EndProcedure
\end{algorithmic}
\columnbreak
\begin{algorithmic}[1]
    \Procedure{PreProcess}{$\mathcal{P},\mathtt{Categories}$}
    \State\Comment{$\mathcal{P}$ is a set of problems}
    \ForAll{$p \in \mathcal{P}$} 
        \State $p.Cat \gets \mathtt{Label}(p,\mathtt{Categories})$
    \EndFor
    \ForAll{$cat \in \mathtt{Categories}$}
        \State $cat.Samples \gets \mathtt{Sample}(\mathcal{P},cat)$
        \State $cat.Ctxt \gets \mathtt{Retriever}(cat.Samples)$
    \EndFor
    \EndProcedure
 \Procedure{PostDisplay}{$p$} 
   \State $\mathcal{F} \gets \{\}$ 
   \ForAll{$\mathtt{model} \in \mathtt{ModelList}$}
     \State $\mathcal{F}.\mathtt{Add}(\Call{Formalize}{p, \mathtt{model}})$
   \EndFor
   \State $\mathtt{summary} \gets \Call{model}{\mathcal{F}, p}$
   \State \Return $\mathcal{F}, \mathtt{summary}$
 \EndProcedure
\end{algorithmic}
    \end{multicols}
    \vspace{-1em}
\end{algorithm}

\ignore{
is this needed?
The \textit{ValidateInLean} and \textit{ParseErrors} functions: Central to our approach is a robust validation system that automatically tests generated Lean 4 code against the formal verification system. The \textit{ValidateInLean} procedure performs syntactic and semantic validation by attempting compilation within a properly configured Lean 4 environment with Mathlib dependencies. This validation occurs in isolated temporary files to prevent interference between different formalizations.
The \textit{ParseErrors} procedure implements intelligent error parsing and categorization, distinguishing between system-level failures (timeouts, missing dependencies, infrastructure errors) and genuine mathematical formalization errors. System failures trigger automatic reprocessing, while mathematical errors are preserved as valuable signal about model performance limitations. This distinction ensures that benchmark statistics reflect actual model capabilities rather than transient technical issues.}
\vspace{-0.3cm}
\paragraph{Step 3: Multi-Model Ensemble and Comparative Analysis.} 

Finally, as outlined in Procedure~$\mathtt{PrepDashboard}$, each problem $p$ is formalized by 12 state-of-the-art models (GPT, Claude, and reasoning variants). This ensemble provides redundancy against model-specific errors, supports comparative evaluation, and increases the likelihood of a correct formalization. The aggregated outputs are then summarized by GPT-5, which ranks models by correctness, completeness, and faithfulness to the problem statement. While this multi-model setup aids evaluation (Section~\ref{sec:autoformalization-evaluation}), the same process can be used for multiple generations of a single model, analogous to self-consistency \citep{wang2023selfconsistencyimproveschainthought}. The comparison also directs users to the correct and incorrect parts of different candidate generations.

\begin{figure}
    \centering
    \includegraphics[width=1\linewidth]{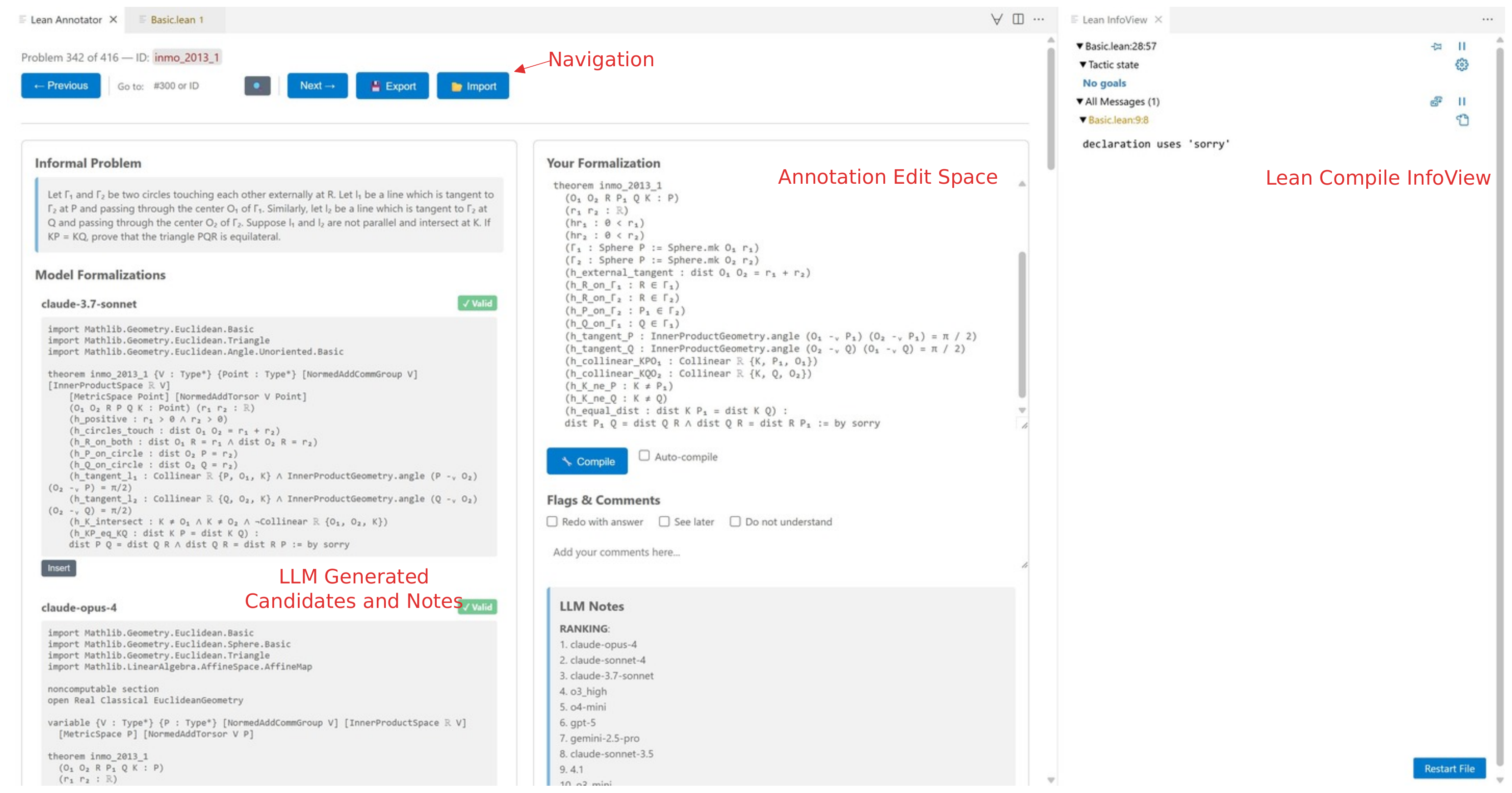}
    \caption{The Annotation Dashboard for the human expert to analyze various ranked options, modify and validate the final entry.}
    \label{fig:dashboard}
\end{figure}

\subsection{Human-AI Collaborative Annotation Dashboard}

We develop a Human–AI collaborative dashboard, as illustrated in Figure \ref{fig:dashboard}, that enables efficient expert review of formalizations generated by LLMs. The interface consolidates all model outputs, validation results, and automated quality assessments into a single interactive workspace, allowing human annotators to efficiently evaluate and refine candidate formalizations.

The dashboard integrates compilation and diagnostic feedback similar to the VS Code Lean extension, allowing experts to compile any candidate directly and view precise error traces. It prioritizes displaying verified results—those that successfully compile and satisfy basic correctness checks—while still providing access to failed attempts for comprehensive analysis.

\textbf{LLM Summaries.} A distinctive feature is the inclusion of AI-generated summaries computed via the $\mathtt{summary}$ variable in Procedure~$\mathtt{PrepDashboard}$. These summaries synthesize insights across model generations, highlighting shared errors, missing conditions, and promising candidates. This guidance allows experts to focus their attention on genuinely ambiguous cases rather than routine validation, significantly reducing cognitive load. Annotators can also merge partial outputs, for example, incorporating a missing condition from one formalization into another, to efficiently produce the final correct formula. Figure \ref{fig:fix_example} shows a direct example of such a speed up.

We release this dashboard as an open-source VS Code extension for broader community use. Beyond the benchmark it can support both informal-to-formal and formal-to-formal annotation tasks across diverse languages, facilitating scalable and high-quality dataset creation.

\textbf{Human Guarantee.} Although the benchmark is constructed through an elaborate AI pipeline, all 312 theorems in \benchmark~are manually verified for both correctness and stylistic quality. Every theorem is rechecked by the other annotator. Annotators observed that many generated formalizations, while valid Lean code, often omit important details of the original problems. Most annotations required editing at varying levels. Section \ref{sec:ann_eff_study} illustrates the necessity of human oversight in creating robust benchmarks, showing an example where a minor error leads to an incorrect autoformalization.

\ignore{
\subsection{Human-AI Collaborative Annotation Dashboard}
\param{this section is too wordy}


Figure \ref{fig:dashboard} illustrates the comprehensive dashboard employed for the efficient human expert review of the final generations from our pipeline. This interface integrates all model outputs, corresponding validation results, automated quality assessments and summaries over the group of formalizations as a whole. The dashboard also incorporates features that the VS Code Lean extension supports, with an option to compile as needed directly. A key feature here is the inclusion of AI-generated annotations, computed via the $\mathtt{summary}$ variable in Procedure~$\mathtt{PrepDashboard}$ -- providing reviewers with comparative analyses of different formalization approaches, identification of common patterns or errors, missed out conditions, and preliminary quality assessments based on compilation success and semantic correctness.
%

These synthesized insights reduce the cognitive load on human annotators, allowing them to maintain the critical human oversight necessary for mathematical accuracy while concentrating their efforts on genuinely ambiguous cases rather than routine validation. Furthermore, the dashboard presents results from different large language models (LLMs), enabling human annotators to consolidate and refine outputs—for instance, by integrating a missing condition from one generated formalization into another. This capability provides human experts with a high-quality initial formulation that can be quickly refined into the final, correct formula, thereby enabling the scalable creation of high-quality formal mathematics benchmarks with minimal human effort.


The dashboard, prioritizes display of verified results—those that successfully compile and pass basic correctness checks—while still providing access to failed attempts and their error traces for comprehensive analysis. This design supports both rapid annotation workflows for ``easy'' cases and detailed investigation for ``hard'' cases of mathematical formalization.

We release this dashboard as a VS Code extension to the community, and it can be generally used for informal to formal, and formal to formal tasks in other languages as well. We hope this extension can serve the wider community create datasets faster.

We provide this dashboard as an easy to use VS Code extension to facilitate broader community adoption.}

\subsection{Annotator efficiency study}
\label{sec:ann_eff_study}
We conduct a limited controlled study with a single annotator to evaluate the impact of LLM-generated candidates and group comparison summaries on informal to formal annotation efficiency for our case. The study compares three workflow settings: 
\begin{enumerate}[left=0pt, itemsep=0em]
    \item \textbf{Full System} provides candidate generations, automated summary notes, and a human-in-the-loop dashboard. This is our proposed system.
    \item \textbf{Masked Candidates} provides the dashboard with all candidate generations, but model identities are hidden and no summary notes are provided, simulating how much help multiple formalizations can give even without having an LLM group critique parts over a manual workflow.
    \item \textbf{Manual Formalization} provides no LLM-generated content a-priori.
\end{enumerate}

We recruited a Lean formalization expert to formalize three sets of 12 RMO problems each (one set from each of three consecutive years, for a total of 36 problems).
This design ensures consistent question styles within each set while providing coverage across problems with similar difficulty.

\begin{wrapfigure}{r}{0.4\textwidth}
    \centering
    \includegraphics[width=0.9\linewidth]{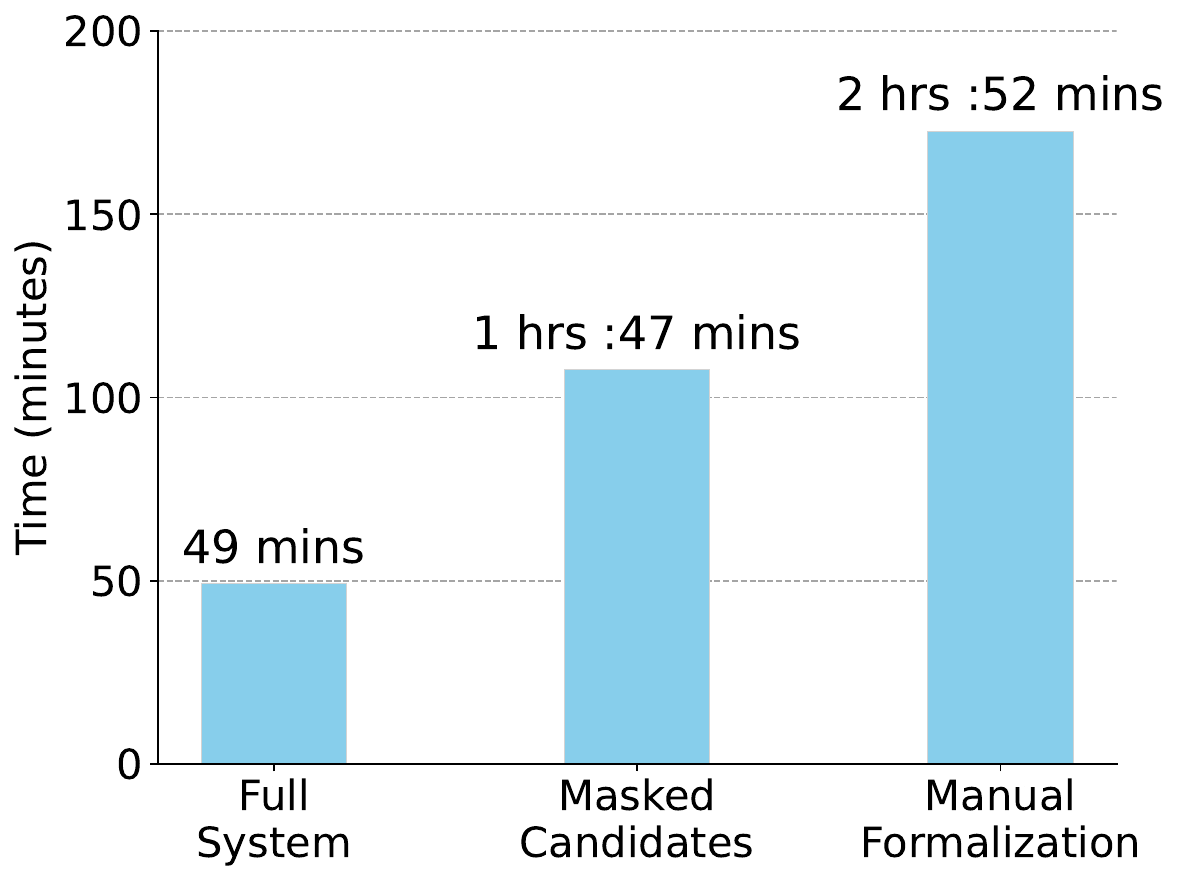}
    \caption{Annotation time for 12 RMO problems under the three workflows}
    \label{fig:time_study}
\end{wrapfigure}

The single annotator formalized three sets across each workflow setting, and is allowed full internet and any AI assistant access throughout every setting. We record the total time spent and the number of problems successfully formalized for comparison. All 36 formalizations were independently double-checked by a second annotator and verified to be correct. 
We present this as a representative mini-study, while acknowledging a potential source of bias arising from the same annotator participating across all three workflows. This observation is preliminary and not tested for statistical significance.

\subsection{Failure Characterization - High precision required in formalizations}

Figure~\ref{fig:time_study} compares annotation times across the three workflow variants. Manual formalization required an average of 14 minutes per problem, comparable to PutnamBench’s 25 minutes and \textsc{miniF2F}’s 10 minutes. Using only multiple masked LLM-generated formalizations reduced this to 9 minutes per problem (a 1.6× speed-up), primarily because the annotator no longer needed to construct problem structures from scratch, and only verify and selectively edit partially correct generations. In contrast, much of the manual effort was spent navigating the mathlib library. With the full system, the average time dropped further to 4 minutes per problem—3.5× faster than manual formalization and 2.2× faster than the multi-generation variant. The annotator noted that when the top-ranked generation in the summary contained errors, the summary usually identified them, and the correct fragment was usually found in another generation. Figure~\ref{fig:fix_example} illustrates this speed-up in practice using a direct case.

We show an example where a small change in the theorem statement renders the problem unprovable. The left was generated by multiple LLMs, and the right is the human-corrected version, using $\mathbb{N}$+ to define positive natural numbers in Lean. 
In classical mathematics, $\mathbb{N}$ typically excludes 0, but in Lean 4 it includes 0. This distinction matters: since if m = n = 0, condition (b) becomes ``0 divides f(0) + f(0)'', which is undefined. Such errors are not caught by the Lean compiler and require careful human annotation.

\begin{mdframed}[linewidth=0.5pt,innerleftmargin=6pt,innerrightmargin=6pt,innertopmargin=6pt,innerbottommargin=6pt]
Let $\mathbb{N}$ denote the set of all natural numbers and let f : $\mathbb{N}$ → $\mathbb{N}$ be a function such that\\
(a) f(mn) = f(m)f(n) for all m,n in $\mathbb{N}$;\\
(b) m + n divides f(m) + f(n) for all m,n in $\mathbb{N}$.\\
Prove that there exists an odd natural number k such that f(n) = n$\wedge$k for all n in $\mathbb{N}$.
\end{mdframed}

\begin{minipage}{0.48\textwidth}
\begin{lstlisting}[style=lean]
import Mathlib

theorem inmo_2018_6 :
  (*$\forall$*) (f : (*$\mathbb{N}$*)+ (*$\rightarrow$*) (*$\mathbb{N}$*)+),
  ((*$\forall$*) m n : 
    (*$\mathbb{N}$*)+, f (m * n) = f m * f n)(*$\rightarrow$*)
  ((*$\forall$*) m n : 
    (*$\mathbb{N}$*)+, (m + n) (*$\mid$*) (f m + f n))(*$\rightarrow$*)
  ((*$\exists$*) k : (*$\mathbb{N}$*), k % 2 = 1 (*$\wedge$*)
  (*$\forall$*) n : (*$\mathbb{N}$*)+, f n = n^k) := by sorry
\end{lstlisting}
\end{minipage}
\hfill
\begin{minipage}{0.48\textwidth}
\begin{lstlisting}[style=lean]
import Mathlib

theorem inmo_2018_6 :
  (*$\forall$*) (f : (*$\mathbb{N}$*) (*$\rightarrow$*) (*$\mathbb{N}$*)),
  ((*$\forall$*) m n : 
    (*$\mathbb{N}$*), f (m * n) = f m * f n)(*$\rightarrow$*)
  ((*$\forall$*) m n : 
    (*$\mathbb{N}$*), (m + n) (*$\mid$*) (f m + f n))(*$\rightarrow$*)
  ((*$\exists$*) k : (*$\mathbb{N}$*), k % 2 = 1 (*$\wedge$*)
  (*$\forall$*) n : (*$\mathbb{N}$*), f n = n^k) := by sorry
\end{lstlisting}
\end{minipage}


\section{Experimental Results}

\subsection{Autoformalization Evaluation}
\label{sec:autoformalization-evaluation}

\paragraph{Setup.} As described in Section \ref{sec:pipeline}, we generate candidate formalizations, given a natural language problem statement, across 12 different general-purpose frontier LLMs. These include Claude Sonnet 4, Claude Opus 4, o3 (high), GPT-5, and Gemini 2.5 Pro, among others. Here, we aim to measure how semantically close the generated formalizations are to the final human annotated formalizations.

\paragraph{Evaluation Metrics.}
Evaluating autoformalization quality presents unique challenges due to the rigorous logical nature of formal mathematical statements, where seemingly minor syntactic variations can alter meaning. To provide a comprehensive assessment, we employ two complementary evaluation metrics that have demonstrated high inter-annotator agreement with human evaluations \citep{liu2025gted}.
\vspace{-0.5em}
\begin{itemize}[left=0pt, itemsep=0em]
\item \textbf{Bidirectional Equivalence (BEq)} \citep{liu2025rethinking} evaluates logical equivalence by attempting to prove each theorem using the other. Given two Lean 4 theorems in \texttt{sorry}-format, \texttt{theorem\_A} and \texttt{theorem\_B}, BEq employs a diverse set of heuristic and LLM-guided tactics to establish proofs in both directions. The formalization is deemed correct only if both directional proofs succeed, ensuring true logical equivalence rather than superficial syntactic similarity.

\item \textbf{Generalized Tree Edit Distance (GTED)} \citep{liu2025gted} measures syntactic similarity by representing Lean theorems as operator trees and computing the normalized cost of transforming the candidate theorem into the human-annotated ground truth. Scores range from 0 to 1 per comparison, with higher values indicating greater structural correspondence.
\end{itemize}

These metrics are contrasting in their approach: BEq captures semantic equivalence through the Lean proof engine's logical reasoning capabilities, while GTED quantifies syntactic structural similarity through tree transformations. This complementary evaluation framework provides both logical validation and structural analysis, offering a complete assessment of autoformalization quality. All implementation details for both the metrics follow the original specifications.

\begin{table}
\begin{minipage}[t]{0.55\textwidth}
\centering
\small
\begin{tabular}{l@{\hskip 4pt}c@{\hskip 4pt}c@{\hskip 4pt}c@{\hskip 4pt}c}
\hline
\textbf{Model} & \textbf{BEq} & \textbf{GTED} & \textbf{\#GTED} & \textbf{Compile} \\
 & & \textbf{Mean} & \textbf{\textgreater0.9} & \textbf{Success} \\
 & \textbf{(312)} & \textbf{(over 312)} & \textbf{(312)} & \textbf{(312)} \\
\hline
Claude Opus 4      & \textbf{54} & \textbf{0.51} & \textbf{138} & \textbf{243} \\
Claude Sonnet 4    & 51 & 0.42 & 103 & 215 \\
Gemini 2.5 Pro     & 47 & 0.24 &  58 & 151 \\
GPT-5              & 38 & 0.48 & 124 & 235 \\
o3 (high)          & 32 & 0.46 & 119 & 205 \\
GPT-4.1                & 29 & 0.39 &  90 & 120 \\
Kimina-Autoformalizer & 17 & 0.19 & 38 & 101\\
Godel-Formalizer-V2 & 52 & 0.47 & 119 & 222\\

\hline
\end{tabular}
\vspace{0.5em}
\caption{Comparing models across BEq, GTED mean, \# of samples with a GTED score \textgreater0.9, and compilation validity counts. GTED scores range from 0-1, higher value meaning better similarity.}
\label{table:autoformalization_eval}
\end{minipage}
\hfill
\begin{minipage}[t]{0.42\textwidth}
\centering
\vspace{-65pt} 
\includegraphics[width=1.0\textwidth]{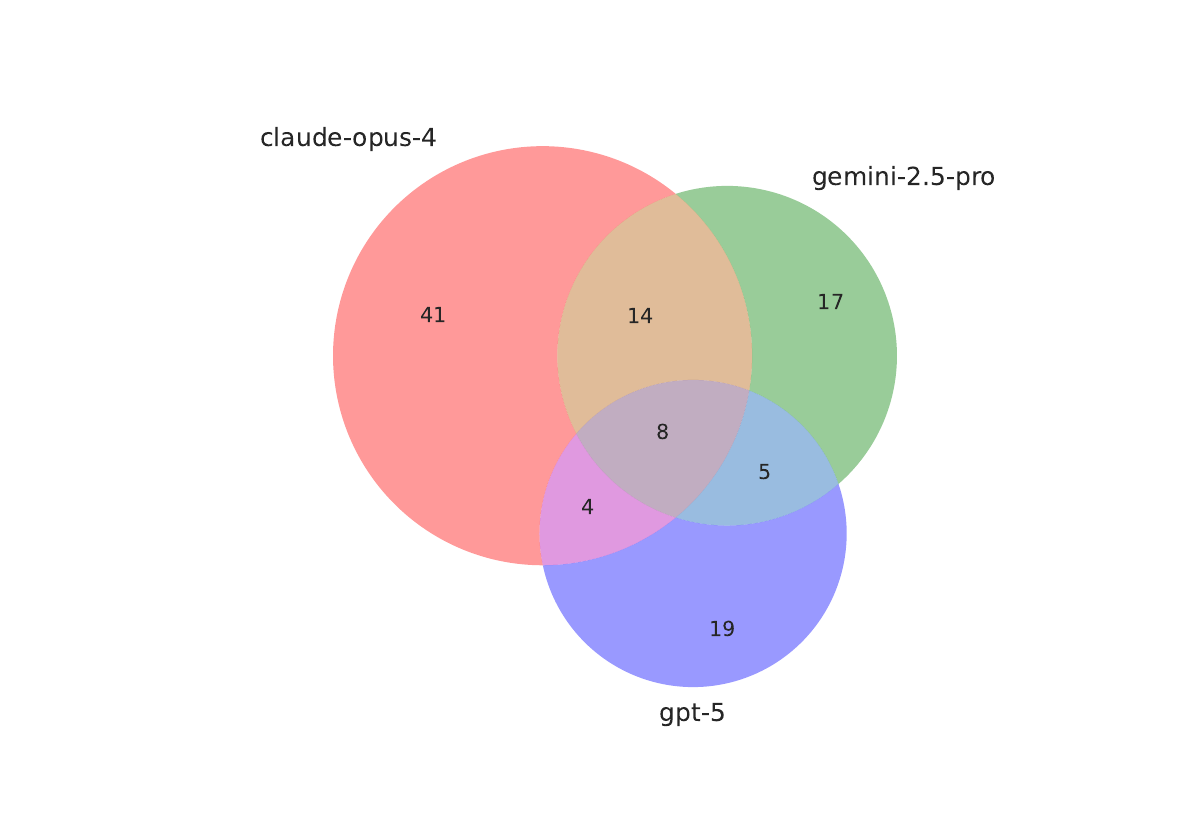}
\captionof{figure}{Venn diagram for BEq-passing problems for three different leading models.}
\label{fig:venn_diagram}
\end{minipage}
\vspace{-5pt}
\end{table}

\paragraph{Results.}
Table \ref{table:autoformalization_eval} depicts the models evaluated against the BEq and GTED metrics. BEq meaures semantic equivalence, while GTED measures syntactic similaritly. Claude Opus 4 does best across metrics, followed by GodelProverV2 8B. Figure \ref{fig:venn_diagram} depicts the overlap on the benchmarks that Claude Opus 4, Gemini 2.5 Pro, and GPT-5 solve (based on BEq metric). This shows that the models have certain complementary abilities and justifies our ensemble approach. In 160 of the 312 problems, at least one generation passed the BEq check, i.e., for 51.2\% of our dataset, an LLM had formalized the problem equivalent to ground truth annotation. Another notable detail is that among Claude Opus 4, Gemini 2.5 Pro, and GPT-5, with a cumulative BEq passing for 108 problems, only 12 ( of 98 total) were from Geometry. This highlights the LLM's difficulty with using Mathlib's lacking support for Olympiad-style geometry.
Appendix \ref{appendix:autoformalization_eval} discusses specific insights from different models.

\subsection{\benchmark~ Technique Evaluation - Ablations} 

\textbf{Ablation for Context and Feedback}: We compare our $\mathtt{Formalize}$ procedure (Algorithm~\ref{alg:main}) against a zero-shot baseline that generates Lean 4 formulas without additional context or iterative feedback. Table~\ref{table:model-wise-lean-validated} reports the percent of problems (312 total) successfully formalized (i.e., passing Lean validation) for both approaches across multiple state-of-the-art language models. We note here that this is only a syntactic notion of theorem correctness, which is necessary, but not sufficient, for full correctness. The results demonstrate that error feedback and documentation retrieval consistently improve success rates across all models. Claude models show particularly notable improvement, starting with low initial compilation success but rapidly refining through available Lean environment knowledge and retrieval context. We use the documentation + feedback generations for our final annotations and evaluations. Additional ablation details are provided in Appendix~\ref{appendix:ablation}.

\ignore{ 
\begin{table}[h!]
\centering
\begin{tabular}{lcccc}
\hline
    & & \textbf{GTED} & \textbf{GTED} & \textbf{Compile} \\
\textbf{Model} & \textbf{BEq} & \textbf{Mean} & \textbf{@0.9} & \textbf{Checked} \\
\hline
claude-opus-4      & 37 & 0.512 & 138 & 243 \\
claude-sonnet-4    & 36 & 0.419 & 103 & 215 \\
gemini-2.5-pro     & 35 & 0.236 &  58 & 151 \\
claude-3.7-sonnet  & 32 & 0.286 &  64 & 171 \\
claude-sonnet-3.5  & 31 & 0.265 &  63 & 162 \\
gpt-5              & 31 & 0.475 & 124 & 235 \\
o4-mini            & 29 & 0.321 &  73 & 165 \\
o3\_high           & 29 & 0.457 & 119 & 205 \\
4.1                & 25 & 0.392 &  90 & 120 \\
o3-mini            & 21 & 0.348 &  84 & 142 \\
4o                 & 10 & 0.267 &  59 & 101 \\
o1-mini            &  7 & 0.336 &  79 &  44 \\
\hline
\end{tabular}
\caption{Comparison of models across BEq, GTED mean, GTED score\textgreater0.9, and sample counts.}
\label{tab:model_comparison_full}
\end{table}

\begin{figure}
    \centering
    \includegraphics[width=0.5\linewidth]{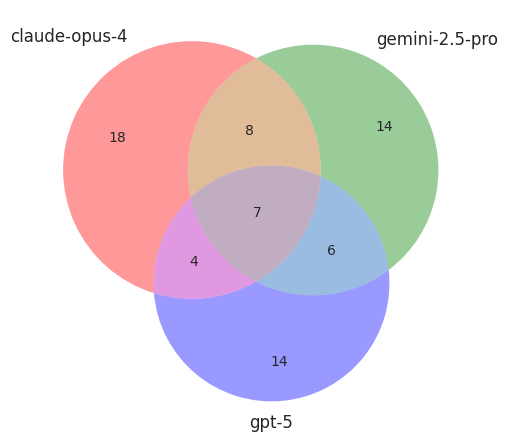}
    \caption{Venn diagram on BEq passing samples.}
    \label{fig:venn_diagram}
\end{figure}

\endignore}




\begin{table}[t]
\centering
\begin{minipage}[t]{0.4\textwidth}
\centering
\small
\begin{tabular}{l@{\hskip 4pt}r@{\hskip 4pt}r}
\hline
\multicolumn{3}{c}{\textbf{Lean Compilation Pass Rate}} \\
\hline
\textbf{Model} & \textbf{ZS (\%)} & \textbf{Doc+FB (\%)} \\
\hline
Claude Opus 4    & 4.1  & \textbf{77.9} \\
GPT-5            & \textbf{30.5} & 75.3 \\
Claude Sonnet 4  & 3.6  & 68.6 \\
o3 (high)        & 22.1  & 65.7 \\
Gemini 2.5 Pro   & 3.8  & 48.4 \\
GPT-4.1          & 13.9  & 38.5 \\
\hline
\end{tabular}
\caption{The percent of Lean-validated formulas generated by the different models in zero-shot (ZS) setting and in the setting with documentation (Doc) and feedback (FB) loops.}
\label{table:model-wise-lean-validated}
\end{minipage}
\hspace{0.02\textwidth}
\begin{minipage}[t]{0.54\textwidth}
\centering
\small
\begin{tabular}{l@{\hskip 4pt}c@{\hskip 4pt}c@{\hskip 4pt}c}
\hline
& \multicolumn{3}{c}{\textbf{Success Rate pass@1}} \\
\cline{2-4}
\textbf{Model} & \textbf{Single Turn} & \textbf{10 Turns} & \textbf{10 Turns} \\
 & \textbf{IMB} & \textbf{IMB} & \textbf{PB} \\
\hline
GPT-4.1         & 0/312 & -  & -  \\
o3 (medium)     & 1/312 & -  & -  \\
Claude Sonnet 4 & 1/312 & 4/312  & -  \\
Gemini 2.5 Pro  & 0/312 & 12/312 & -  \\
GPT-5           & 1/312 & \textbf{36}/312 & 28/660 \\
DeepSeekProver V2 7B & 3/312 & 24/312 & - \\
GodelProver V2 8B  & \textbf{7}/312 & \textbf{36}/312 & - \\
\hline
\end{tabular}
\caption{Success Rates, all pass@1, of various frontier models on \benchmark. Success Rates here refer to Lean Verifiable proofs submitted by the models. IMB: \benchmark, PB: PutnamBench}
\label{table:atp_results}
\end{minipage}\vspace{-18pt}
\end{table}


\subsection{Automated Theorem Proving Evaluation}
We measure the difficulty \benchmark~ poses to frontier general purpose LLMs for formal theorem proving, by testing them for completing the theorem statement with a valid proof.
We evaluate success rate in single-turn and 10-turn settings with Lean compiler feedback. All results are reported on pass@1 metric.
We carry out these evaluations across the best frontier models available to us, Claude Sonnet 4, GPT-5, GodelProverV2, Gemini 2.5 Pro, and others at default parameters. 
In single turn mode, only one problem was solved (inmo\_2015\_5, requiring Mathlib's pitot\_theorem) by Claude Sonnet 4, GPT-5, and o3 (medium) each, across the 312 strong \benchmark.
With 10 turns and iterative refinement in a ReAct \citep{yao2023reactsynergizingreasoningacting} like procedure, GPT-5 achieves 36/312 (11\%) on \benchmark~and 28/660 (4\%) on PutnamBench, demonstrating comparable difficulty between the two benchmarks. Notably, no geometry problems were solved by any model. The multi-turn construction used here follows chain-of-thought based practices established in prior systems~\citep{chen2025seedproverdeepbroadreasoning, shen2025realproverretrievalaugmentedlean, ji2025leanabellproverv2verifierintegratedreasoningformal}. Detailed setup and results are described in Appendix \ref{appendix:atp}.



\section{Conclusion}




We introduce \benchmark, a new benchmark of human-verified Lean 4 formalizations for Olympiad-level mathematics problems. The dataset is designed to evaluate the limits of large language models in formal reasoning, where our comparative analysis shows that even frontier models solve only a single problem in single-turn settings and reach just 11\% success with iterative refinement. These results highlight the substantial gap between current LLM capabilities and the rigor required for formal mathematics, and the dataset complexity.
To support progress in this direction, we further present a structured annotation framework that leverages documentation retrieval, compiler feedback, and multi-model aggregation to streamline manual formalization. We release both the dataset and an integrated VS Code dashboard extension as open resources to advance research in neural theorem proving and autoformalization.


\bibliography{iclr2026_conference}
\bibliographystyle{iclr2026_conference}

\appendix
\newpage
\section{\benchmark}

\textbf{Problem Domains.} 
\benchmark~ problems are traditionally sourced from a fixed set of topics: algebra, euclidean geometry, elementary number theory, and combinatorics. Calculus, set theory, and linear algebra are not part of the official syllabus. This restriction makes the benchmark more uniform in scope, while still allowing for deeper exploration of problem-solving within each domain. However, the problems—especially from INMO—demonstrate remarkable internal diversity and depth. 
For instance, geometry problems frequently involve classical triangle centers (e.g., orthocenter, centroid) or cyclic quadrilaterals, which often demand nontrivial formalization in Lean and combine diagrammatic insight with auxiliary constructions. Similarly, number-theoretic questions tend to hinge on clever use of parity, bounding, or invariant arguments, reflecting the benchmark’s emphasis on depth rather than breadth. Set Theory and Combinatorics are treated as a single domain due to the substantial overlap observed in the corresponding problems.

\textbf{Underrepresented Domains.} Domains like geometry, combinatorics, and counting are under-formalized even in larger datasets \citep{zheng2022minif2f, tsoukalas2024putnambench, song2025leangeoformalizingcompetitionalgeometry, yu2025formalmathbenchmarkingformalmathematical, ren2025deepseekproverv2advancingformalmathematical}. Another reason for selecting RMO and INMO for formalization was their unique syllabus and domain compared to other exams around covered in literature.

\textbf{RMO and INMO exams.} The Regional Mathematics Olympiad and the Indian National Mathematics Olympiad exams are held annually, and are used to select India's most promising high-school student. Students who pass the RMO qualify to take the INMO, which is a significantly more difficult national-level test. Students typically need to pass another exam before qualifying for RMO. Around 7000 students qualify to write the RMO, and 1000 to write the INMO every year from all over India. 

\subsection{Example formalizations from \benchmark}

\subsubsection{Geometry problem formalization}
Consider the following problem statement:
\begin{mdframed}[linewidth=0.5pt,innerleftmargin=6pt,innerrightmargin=6pt,innertopmargin=6pt,innerbottommargin=6pt]
In a triangle ABC, let D be a point on the segment BC such that AB + BD = AC + CD. Suppose that the points B, C and the centroids of triangles ABD and ACD lie on a circle. Prove that AB = AC.
\end{mdframed}

Its formalization in \benchmark:

\begin{lstlisting}[style=lean]
variable {V : Type*} {P : Type*} [NormedAddCommGroup V] 
  [InnerProductSpace (*$\mathbb{R}$*) V] [MetricSpace P] [NormedAddTorsor V P]

theorem inmo_2014_1 (A B C D : P) 
  (hD : (*$\exists$*) t : (*$\mathbb{R},\;  0 \leq t \; \wedge \; t \leq 1 \; \wedge \; D =$*) AffineMap.lineMap B C t)
  (h_sum : dist A B + dist B D = dist A C + dist C D)
  (h_concyclic : Concyclic {B, C, 
    centroid (*$\mathbb{R}$*) {A, B, D} id, 
    centroid (*$\mathbb{R}$*) {A, C, D} id}) :
  dist A B = dist A C := by sorry
\end{lstlisting}

\ignore{ 
As a second example, consider the following problem:
\begin{mdframed}[linewidth=0.5pt,innerleftmargin=6pt,innerrightmargin=6pt,innertopmargin=6pt,innerbottommargin=6pt]
Let $n$ be a natural number. Prove that:
\[ \left\lfloor \frac{n}{1} \right\rfloor + \left\lfloor \frac{n}{2} \right\rfloor + \left\lfloor \frac{n}{3} \right\rfloor + \cdots + \left\lfloor \frac{n}{n} \right\rfloor + \left\lfloor \sqrt{n} \right\rfloor \]
is even.
    theorem inmo_2014_2 (n : *$\mathbb{N}$*) : Even ((Finset.sum (Finset.range n) fun i => ⌊(n : ℝ) / ((i + 1) : ℝ)⌋) + ⌊Real.sqrt n⌋) := by sorry
\end{mdframed}



theorem inmo_2014_2 (n : ℕ) : Even ((Finset.sum (Finset.range n) fun i => ⌊(n : ℝ) / ((i + 1) : ℝ)⌋) + ⌊Real.sqrt n⌋) := by sorry

\endignore}

\subsubsection{Solve type problem formalization}

There are other methods of formalization which do not include explicitly mentioning the final answer in case of solve type problems \citep{tsoukalas2024putnambench}, however we only support the question conversion type formalization in this release.

Consider the following problem description.
\begin{mdframed}[linewidth=0.5pt,innerleftmargin=6pt,innerrightmargin=6pt,innertopmargin=6pt,innerbottommargin=6pt]
Suppose $n \geq 0$ is an integer and all the roots 
    of $x^3 + \alpha x + 4 - (2 \cdot 2016^n) = 0$ are integers. Find all possible values of $\alpha$.
\end{mdframed}
This is a problem that requires a solution. We assume a solution is given.
For such cases, the formula states that the given solution is actually 
correct.

\begin{lstlisting}[style=lean]
import Mathlib.Data.Int.Basic
import Mathlib.Algebra.Polynomial.Basic

open Polynomial

theorem inmo_2017_2 (n : (*$\mathbb{N}$*) ) ( (*$\alpha$*) : (*$\mathbb{Z}$*) ) :
    ( (*$\exists$*) u v w : (*$\mathbb{Z}$*) ,
      (X ^ 3 + C (*$\alpha$*) * X + C (4 - 2 * (2016 ^ n))
        = (X - C u) * (X - C v) * (X - C w)) (*$\rightarrow$*) (*$\alpha$*) = (-3 : (*$\mathbb{Z}$*) ) := by
  sorry
\end{lstlisting}

\ignore{
\subsubsection{High precision required in formalizations}
\label{appendix:simple_differences}

We list an example formalization where small differences in the theorem statement make the entire problem unprovable. The left was generated by multiple LLMs, and the right was the human annotation after making the correction of using $\mathbb{N}$+ to define positive natural numbers in Lean. 

Natural numbers in the classical (informal) mathematical notation are denoted by the symbol $\mathbb{N}$. Usually, and in the context of this question, 0 is not included. However, in Lean 4, $\mathbb{N}$ denotes the set of natural numbers that begin with 0. This distinction plays an important role, since if m = n = 0, condition (b) becomes ``0 divides f(0) + f(0)'', which is undefined.

Such an error would not be bought up by the lean compiler and could slip by easily without careful annotations.

\begin{mdframed}[linewidth=0.5pt,innerleftmargin=6pt,innerrightmargin=6pt,innertopmargin=6pt,innerbottommargin=6pt]
Let $\mathbb{N}$ denote the set of all natural numbers and let f : $\mathbb{N}$ → $\mathbb{N}$ be a function such that\\
(a) f(mn) = f(m)f(n) for all m,n in $\mathbb{N}$;\\
(b) m + n divides f(m) + f(n) for all m,n in $\mathbb{N}$.\\
Prove that there exists an odd natural number k such that f(n) = n$\wedge$k for all n in $\mathbb{N}$.
\end{mdframed}

\begin{minipage}{0.48\textwidth}
\begin{lstlisting}[style=lean]
import Mathlib

theorem inmo_2018_6 :
  (*$\forall$*) (f : (*$\mathbb{N}$*)+ (*$\rightarrow$*) (*$\mathbb{N}$*)+),
  ((*$\forall$*) m n : 
    (*$\mathbb{N}$*)+, f (m * n) = f m * f n)(*$\rightarrow$*)
  ((*$\forall$*) m n : 
    (*$\mathbb{N}$*)+, (m + n) (*$\mid$*) (f m + f n))(*$\rightarrow$*)
  ((*$\exists$*) k : (*$\mathbb{N}$*), k % 2 = 1 (*$\wedge$*)
  (*$\forall$*) n : (*$\mathbb{N}$*)+, f n = n^k) := by sorry
\end{lstlisting}
\end{minipage}
\hfill
\begin{minipage}{0.48\textwidth}
\begin{lstlisting}[style=lean]
import Mathlib

theorem inmo_2018_6 :
  (*$\forall$*) (f : (*$\mathbb{N}$*) (*$\rightarrow$*) (*$\mathbb{N}$*)),
  ((*$\forall$*) m n : 
    (*$\mathbb{N}$*), f (m * n) = f m * f n)(*$\rightarrow$*)
  ((*$\forall$*) m n : 
    (*$\mathbb{N}$*), (m + n) (*$\mid$*) (f m + f n))(*$\rightarrow$*)
  ((*$\exists$*) k : (*$\mathbb{N}$*), k % 2 = 1 (*$\wedge$*)
  (*$\forall$*) n : (*$\mathbb{N}$*), f n = n^k) := by sorry
\end{lstlisting}
\end{minipage}}

\section{Autoformalization Evaluation}
\label{appendix:autoformalization_eval}

\begin{figure}[htbp]
    \centering
    \includegraphics[width=0.75\textwidth]{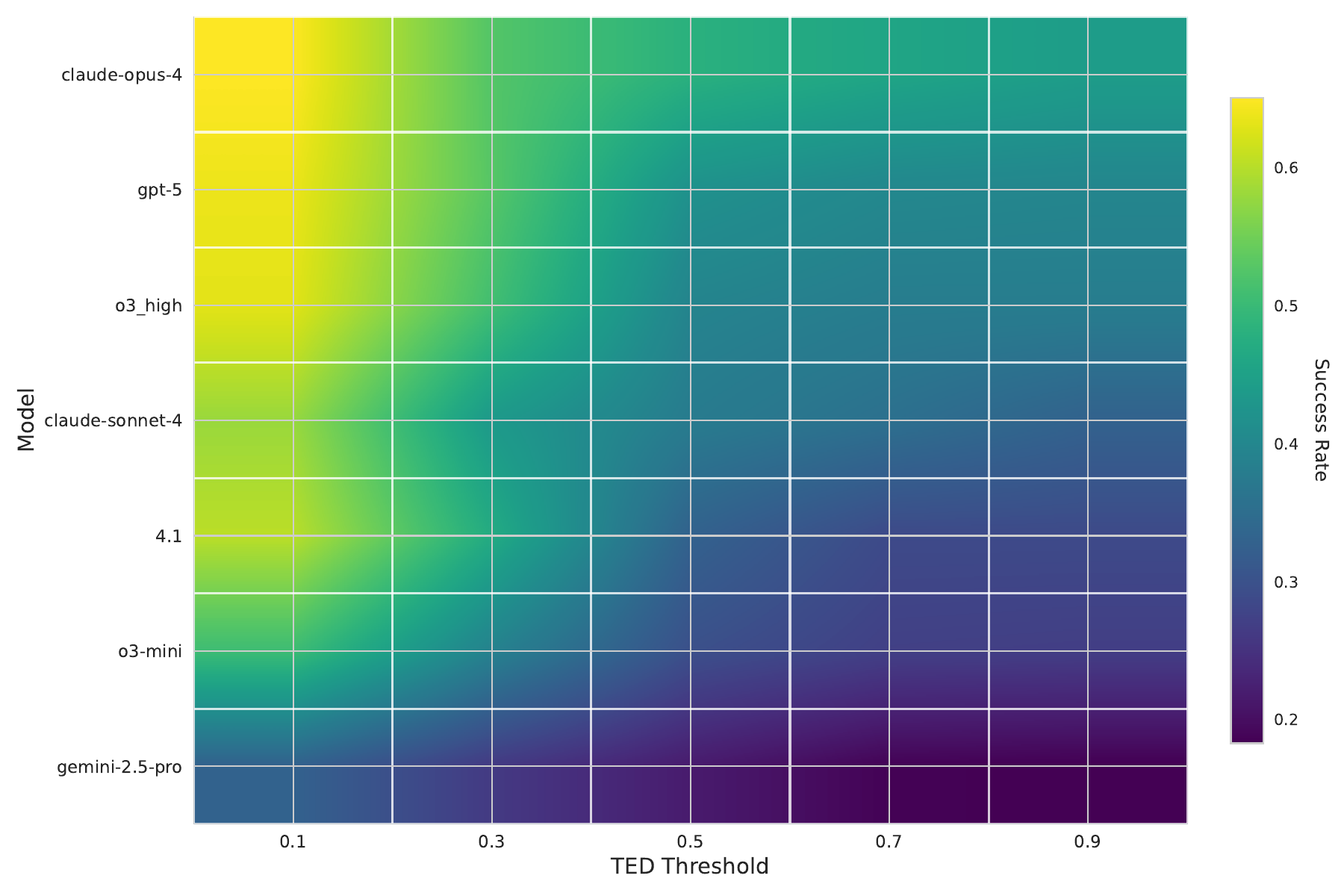}  
    \caption{Heatmap for GTED based on thresholds. Better models see a lighter hue.}
    \label{fig:heatmap}
\end{figure}

Table \ref{table:autoformalization_eval_comparison_full} shows the BEq and GTED scores for 14 models. Figure \ref{fig:heatmap} shows the heatmap among different models (y-axis) with respect to their GTED scores across the 312 comparisons, with a lighter overall hue of what ratio of comparisons pass various thresholds (x-axis). We also run BEq using a general purpose model (Claude Sonnet 4) as the underlying prover, and the trends remain the same.

\begin{table}[ht]
\centering
\begin{tabular}{lccccc}
\hline
\textbf{Model} & \textbf{BEq} & \textbf{BEq} & \textbf{GTED} & \textbf{\#GTED} & \textbf{Compile} \\
 & InternLM2-7B@16& Claude Sonnet 4@1 & \textbf{Mean} & \textbf{\textgreater0.9} & \textbf{Success} \\
 & \textbf{(312)} & \textbf{(312)} & \textbf{(312)} & \textbf{(312)} \\
\hline
Claude Opus 4      & \textbf{54} & \textbf{67} & \textbf{0.51} & \textbf{138} & \textbf{243} \\
Claude Sonnet 4    & 51 & 54 & 0.42 & 103 & 215 \\
Godel-Formalizer-V2& 52 & -  & 0.47 & 119 & 222 \\
Gemini 2.5 Pro     & 47 & 44 & 0.24 &  58 & 151 \\
GPT-5              & 38 & 36 & 0.48 & 124 & 235 \\
o3\_high           & 32 & 30 & 0.46 & 119 & 205 \\
GPT-4.1            & 29 & 33 & 0.39 &  90 & 120 \\
Kimina-Autoformalizer& 17 & -& 0.19 &  38 & 101 \\
Claude Sonnet 3.7  & -  & 37 & 0.29 &  64 & 171 \\
Claude Sonnet 3.5  & -  & 36 & 0.27 &  63 & 162 \\
o4-mini            & -  & 30 & 0.32 &  73 & 165 \\
o3-mini            & -  & 28 & 0.35 &  84 & 142 \\
GPT-4o             & -  & 14 & 0.27 &  59 & 101 \\
o1-mini            & -  &  9 & 0.34 &  79 &  44 \\
\hline
\end{tabular}
\vspace{0.5em}
\caption{Comparing all 12 models across BEq, GTED mean, \# of samples with a GTED score \textgreater0.9, and compilation validity counts.}
\label{table:autoformalization_eval_comparison_full}
\end{table}

\subsection{Model failure modes in Autoformalization}

Based on empirical evaluation using \benchmark, a benchmark of 312 formally verified Lean4 theorems from Indian Mathematical Olympiad problems, we analyze systematic failure patterns across major model families in autoformalization tasks.

\subsubsection{Anthropic Claude Model Family}

The Claude 4 series demonstrates the strongest overall performance in autoformalization metrics, with Claude Opus 4 achieving the highest BEq score (67/312) and compilation success rate (243/312).

\paragraph{Claude Opus 4.}
Claude Opus 4 leads across multiple evaluation metrics with 67 BEq-passing formalizations and 0.512 mean GTED score, representing the current state-of-the-art in mathematical autoformalization. Characteristic strengths and limitations:
\begin{itemize}[left=0pt, itemsep=0em]
    \item \textit{Superior Semantic Accuracy}: Achieves highest BEq logical equivalence with ground truth formalizations
    \item \textit{Compilation Reliability}: 77.9\% of generations compile successfully in Lean4 
    \item \textit{Iterative Refinement Capability}: Shows dramatic improvement from 17 to 310 successful compilations when provided with error feedback and knowledge base access
    \item \textit{Domain-Specific Challenges}: Particularly struggles with geometry problems, contributing only 6 of 67 successful formalizations
\end{itemize}

Sample Claude Opus 4 autoformalization which is also semantically correct:
\begin{lstlisting}[style=lean]
theorem inmo_2014_2 (n : (*$\mathbb{N}$*)) :
  Even ((Finset.sum (Finset.range n)
    fun i => (*$\lfloor$*)(n : (*$\mathbb{R}$*)) / ((i + 1) : (*$\mathbb{R}$*))(*$\rfloor$*)) + (*$\lfloor$*)Real.sqrt n(*$\rfloor$*)) := by
sorry
\end{lstlisting}

\paragraph{Claude Sonnet 4.}
Demonstrates consistent performance with 54 BEq successes and more conservative formalization approaches, showing strong baseline capabilities with room for refinement. Characteristic strengths and limitations:
\begin{itemize}[left=0pt, itemsep=0em]
    \item \textit{Strong Improvement with Support}: Only 15 successful compilations without assistance, increases to 272 successful compilations with knowledge base and feedback
    \item \textit{Structural Similarity}: Achieves 0.419 mean GTED score, indicating reasonable syntactic accuracy
\end{itemize}


\subsection{OpenAI GPT Model Family}

The GPT series shows varied performance across different model variants, with GPT-5 achieving moderate success but significant challenges in zero-shot scenarios.

\paragraph{GPT-5.}
Despite being a one of the best general purpose model, GPT-5 achieves only 36 BEq successes, demonstrating the complexity of mathematical formalization tasks. Performance analysis:
\begin{itemize}[left=0pt, itemsep=0em]
    \item \textit{Strong Zero-Shot Compilation}: 127 successful compilations without assistance, highest among all models
    \item \textit{Knowledge Base Responsiveness}: Improves to 297 successful compilations with documentation and feedback
    \item \textit{Semantic Gaps}: Despite compilation success, only 11.5\% achieve BEq semantic equivalence with ground truth
    \item \textit{High Structural Similarity}: 0.475 mean GTED score suggests good syntactic understanding
\end{itemize}

\paragraph{GPT-4.1.}
Shows the weakest overall performance among evaluated models, particularly in zero-shot scenarios. Performance analysis:
\begin{itemize}[left=0pt, itemsep=0em]
    \item \textit{Poor Zero-Shot Performance}: Only 58 successful compilations without assistance
    \item \textit{Limited Semantic Understanding}: 33 BEq successes despite reasonable compilation rates
    \item \textit{Lowest Mean GTED}: 0.392 structural similarity score indicates syntactic challenges
\end{itemize}

\subsection{OpenAI o-Series Models}

The o3 reasoning model demonstrates interesting patterns, showing strong zero-shot compilation but moderate semantic accuracy.

\paragraph{o3 (High Reasoning).}
Achieves balanced performance with distinctive reasoning-based approach to formalization. Performance analysis:
\begin{itemize}[left=0pt, itemsep=0em]
    \item \textit{Strong Zero-Shot Reasoning}: 92 successful compilations without documentation
    \item \textit{Improvement with Context}: Increases to 263 successful compilations with knowledge base
    \item \textit{Semantic Accuracy Challenges}: Only 30 BEq successes despite compilation strength
    \item \textit{Reasoning Overhead}: Sometimes refuses to formalize when uncertain, as evidenced by responses stating inability to find solutions
\end{itemize}

\subsection{Google Gemini}

Gemini 2.5 Pro shows the weakest performance among frontier models, despite known for being very good at informal mathematical reasoning \citep{huang2025winninggoldimo2025, varambally2025hilbertrecursivelybuildingformal}, particularly struggling with Lean4-specific syntax and mathematical library integration.

\paragraph{Gemini 2.5 Pro}
Demonstrates significant challenges across all evaluation metrics with only 44 BEq successes. Performance analysis:
\begin{itemize}[left=0pt, itemsep=0em]
    \item \textit{Poor Zero-Shot Performance}: Only 16 successful compilations without assistance
    \item \textit{Limited Improvement Capacity}: Increases to only 184 compilations with full support system
    \item \textit{Lowest Semantic Accuracy}: 14.1\% BEq success rate among successful compilations
    \item \textit{Structural Understanding Gaps}: 0.236 mean GTED score indicates fundamental syntactic challenges
\end{itemize}

Gemini 2.5 Pro's performance highlights how mathematical reasoning and formal notation in Lean are two different tasks.

\subsection{Cross-Family Universal Patterns}

\subsubsection{Zero-Shot vs. Assisted Performance Gap}
All model families show dramatic performance improvements when provided with domain-specific knowledge base and iterative error feedback, with improvement ratios ranging from 2.3x (GPT-5) to 18.2x (Claude Opus 4).

\subsubsection{Theorem Proving vs. Formalization Gap}
Despite varying formalization capabilities, all models achieve nearly identical theorem proving performance on single turn, with only Claude Sonnet 4, GPT-5, and O3 (medium) successfully proving a single theorem (inmo\_2015\_5) that relies on existing Mathlib theorems. However, over multiturn, GPT-5 gives performance better than many finetuned (for Lean notation) models over PutnamBench \citep{lin2025goedelproverfrontiermodelopensource, wang2025kiminaproverpreviewlargeformal}.

\subsubsection{Geometry Domain Challenges}
Across all model families, geometry problems present the greatest formalization challenge. Among 108 successful BEq formalizations from the top three models (Claude Opus 4, Gemini 2.5 Pro, GPT-5), only 12 originated from geometry problems, indicating systematic difficulties with spatial reasoning and coordinate-free geometric formalization.

\subsubsection{Compilation vs. Semantic Correctness}
A consistent pattern emerges where compilation success significantly exceeds semantic correctness across all families. Claude Opus 4's 77.9\% compilation rate yields only 21.5\% semantic accuracy, highlighting the challenge of generating syntactically valid but semantically incorrect formalizations. Example of a half incomplete formalization that passes Lean compiler:

\begin{lstlisting}[style=lean]
-- Typical geometry problem requiring complex formalization:
theorem triangle_centroid_property (A B C D : P)
  (hD : (*$\exists$*) t : (*$\mathbb{R}$*), 0 (*$\leq$*) t (*$\wedge$*) t (*$\leq$*) 1 (*$\wedge$*) D = AffineMap.lineMap B C t)
  (h_sum : dist A B + dist B D = dist A C + dist C D)
  (h_concyclic : Concyclic {B, C,
    centroid (*$\mathbb{R}$*) {A, B, D} id,
    centroid (*$\mathbb{R}$*) {A, C, D} id}) :
  dist A B = dist A C := by sorry
-- All models struggle with geometric abstractions like centroids and
concyclic points
\end{lstlisting}

\subsection{Implications for Mathematical AI Development}

The empirical findings reveal that current autoformalization capabilities remain limited despite advances in general reasoning. The consistent pattern of strong compilation performance coupled with weak semantic accuracy suggests that models can learn Lean4 syntax but struggle with mathematical meaning preservation. The dramatic improvement from iterative feedback indicates that human-AI collaborative approaches may be more promising than fully automated formalization for complex mathematical domains.
\section{Experimental Setup}
\label{appendix:setup}

All formalizations and experiments were conducted using the Lean theorem prover, version 4.22.  
We relied on the \texttt{mathlib} library at commit \texttt{f858fcc3b49c546705ba7d79c58217e85aaa5f0e} to ensure reproducibility and consistency across our proofs and auxiliary results.

Our computational environment consisted of:
\begin{itemize}
    \item \textbf{Hardware:} 8-core CPU, 32~GB RAM
    \item \textbf{OS:} Windows 11 Pro (64-bit)
    \item \textbf{Lean Toolchain:} Installed via \texttt{elan}, with \texttt{lake} for project management
\end{itemize}

All proofs were compiled and verified using Lean’s native \texttt{lake build} system without additional modifications to \texttt{mathlib}.  
The experiment scripts, proof files, and configuration details are provided in the supplementary material to facilitate full reproducibility.

\subsection{Impact of LLM Notes}

\begin{figure}[t]
    \centering
    \includegraphics[width=\linewidth]{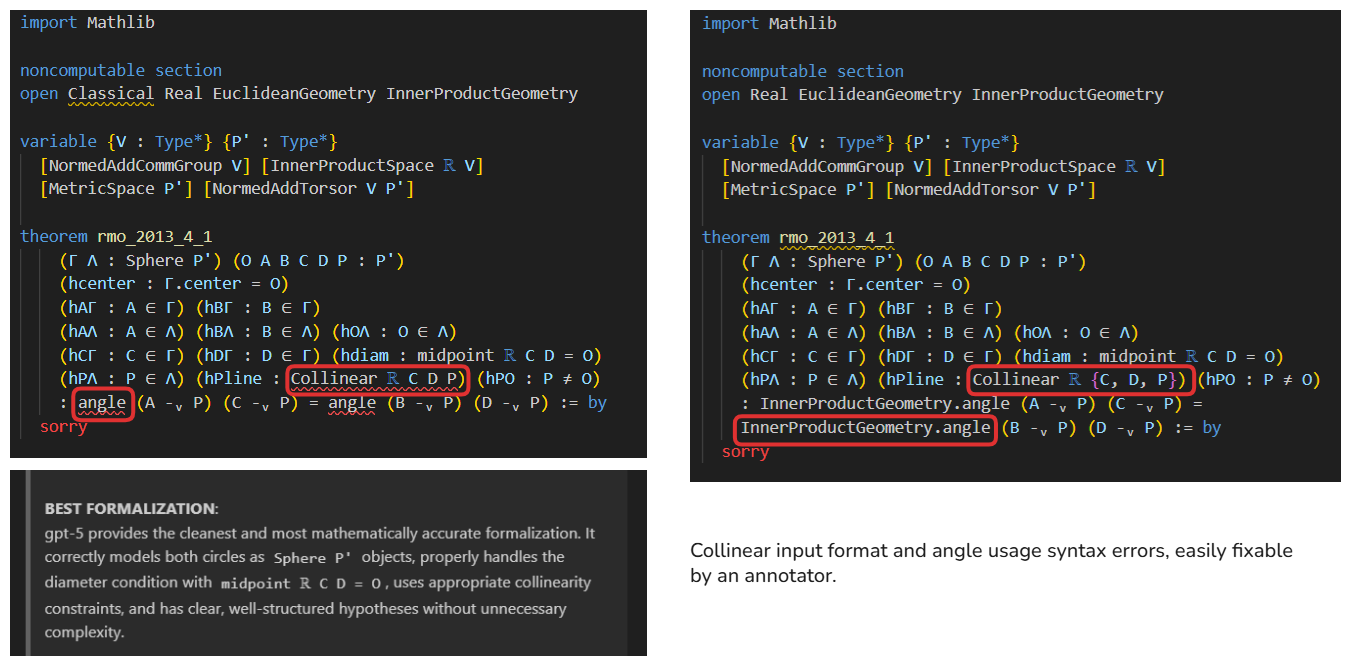}
    \caption{Minimal edits needed to fix a broken, but mathematically sound generation, as pointed out by the LLM summary. Reducing human effort by a margin.}
    \label{fig:fix_example}
\end{figure}

In this section we discuss the impact of LLM generated notes to reduce effort of annotation. 
While our primary focus is on improving proof generation through activation steering, we observe an additional benefit: the method can substantially reduce the human effort required for data annotation and theorem formalization. Figure~\ref{fig:fix_example} illustrates a representative example where steering produces mathematically sound code that requires only minimal corrections to become a valid formal proof.

In this example, the model generates a Lean theorem about geometric relationships in Euclidean space. The initial generation contains two primary issues: incorrect collinear input formatting and improper angle usage syntax. However, the mathematical reasoning underlying the proof is correct, as confirmed by the model's own natural language summary stating that ``gpt-5 provides the cleanest and most mathematically accurate formalization.'' The required fixes are straightforward syntactic corrections that can be applied by domain experts with minimal effort.

This pattern suggests that activation steering not only improves proof success rates but also generates ``near-miss'' proofs that are closer to correctness than baseline outputs. Rather than producing completely invalid formal statements, steered models tend to generate mathematically coherent structures with localized syntax errors. This property significantly reduces the annotation burden for creating training datasets, as human experts can focus on lightweight editing rather than complete theorem reconstruction.

The implications for scalable dataset creation are substantial. Traditional approaches to building formal mathematics datasets require expert mathematicians to write complete formalizations from scratch---a time-intensive process that limits dataset scale. Our approach enables a more efficient workflow: models generate candidate formalizations that capture the essential mathematical content, while human experts provide targeted corrections to syntax and edge cases. This collaborative paradigm could accelerate the development of large-scale formal mathematics corpora needed to train more capable theorem-proving systems.

We note that this benefit emerges naturally from our steering methodology rather than being explicitly optimized for. The fact that informal reasoning guidance leads to more structured, correctable outputs suggests that the underlying activation patterns encode not just proof search strategies, but also adherence to formal syntax conventions. This observation warrants further investigation in future work focused specifically on human-AI collaborative formalization workflows.

\subsection{Knowledge Base Learning Prompt}
\label{prompt:kb}

\subsubsection{System Message}
\label{prompt:kb_system}

\begin{lstlisting}[style=promptstyle]
You are a mathematical documentation agent specializing in the Lean 4 Mathlib library. Your job is to explore the Mathlib repository and create a concise, practical documentation summary focused on mathematical formalization and theorem proving. You have been given access to a yet unreleased version of this library, which you must go through and pick out all relevant imports based on the type of problem the user is trying to solve. The repository contains a comprehensive library of formalized mathematics for Lean 4.
The repository will have file names and folder names representative of its content.

Your every response must be a tool call.

The documentation will be used by new lean users, who will use it as a guide to write all their imports, writing notations and rely solely on it to make the correct imports.

WORKFLOW:
1. Use run_bash to explore the repository (ls, cd, cat, grep, find, etc.)
2. Take notes by writing to files in {working_directory}. You are currently at this directory. Please do not make any changes outside of this directory, or delete any existing file.
    i.      First read all the given examples, and create a list keywords, such that each keyword is a concept that appears in any question.
    ii.     Keywords should also include common patterns like how to express "point lies on line segment", "lines are parallel/perpendicular", ratios and divisions of segments.
    iii.    Add these keywords to your notes file, so you can refer to them for completion later on.
    iv.     Understand the kind of problems the documentation needs to deal with, and select what goes in accordingly.
3. When you have sufficient information, use final_submit with a complete documentation string

EXPLORATION STRATEGY:
- Examine the main mathematical domains asked by the user
- Look for key theorem statements and their dependencies
- Pay attention to naming conventions and mathematical abstractions
- Use the given sample of examples to understand what parts to focus on
- Look for file names, folder names, documentation, examples, source code to know their subject
- Focus on user-facing functionality
- Use {working_directory} for any notes (absolute paths since you'll be changing directories)
- You decide when you have enough information to create the final documentation

FINAL DOCUMENTATION FORMAT:
Organize your final output into exactly these 4 sections:

## 1. Installation & Import
- How different imports are situated in the mathlib file hierarchy
- Essential import statements for different mathematical domains
- Any setup requirements, like opening some namespace for certain symbols, literals, notations or declarations.

## 2. Available Namepaces and Symbols
- Group related functionality together
- Since you will be given a field by the user, focus only on that and related thing you see in the examples
- Important theorem statements in each subdomain
- Common mathematical objects and their properties
- Exhaustive list of all the functions avaliable for use

## 3. Minimal Usage Example  
- Simple theorem statement (with sorry, ignore proofs)
- Basic mathematical definitions
- Make some imports, and open some namespaces and scopes
- All sample codes **must** be complete and well explained, or else it can confuse the readers on what a complete theorem code looks like
- Do not leave parts of example code as comments
- Give examples for the kind of stuff the reader will be dealing with when trying to formalize the problem statement
- Lean has difficult type setups, so be sure to explain those with examples
- Should work out of the box

## 4. Common Pitfalls & Gotchas
- Common mistakes when formalizing mathematics
- Type class resolution issues  
- Mathematical notation vs. Lean syntax differences

## 5. Key Files Structure
- An ascii directory tree of all the important/related files and packages

If some concept appears even once in the examples, make sure to cover that in your documentation. It should be **complete**, don't skip concepts randomly.
Do not be afraid to make long if it needs to be.

Remember: Your goal is to create a practical cheat sheet that gets developers productive quickly. It is okay if its long as long as we are putting relevant information and are correct.
\end{lstlisting}

\subsubsection{User Message}
\label{prompt:kb_user}

\begin{lstlisting}[style=promptstyle]
Problem Description: I want to understand what all library modules are available to me for autoformalizing **Set Theory & Combinatorics** olympiad like problem statements into lean 4. I only care about autoformalizing the theorem part, so things like tactics and everything related to solving the problem are unnecessary. Only things relevant to the theorem statement are useful. I am interested in:
- All the necessary and relevant imports, their correct paths
- How to open the correct namespace or scope to use particular symbols or literals in lean
- Examples of using them
- Other things to note
I'll attach some examples of the type of questions I am trying to write as a lean theorem.

Examples: Samples of the kind of questions whose autoformalization I'll be doing:
- All the 7-digit numbers containing each of the digits 1, 2, 3, 4, 5, 6, 7 exactly once, and not divisible by 5, are arranged in the increasing order. Find the 2000-th number in this list.
- Prove that the number of triples $(A, B, C)$ where $A, B, C$ are subsets of $\{1, 2, \ldots , n\}$ such that $A \cap B \cap C = \emptyset$, $A\cap B \neq \emptyset$, $B\cap C\neq \emptyset$ is $7^n - 2\cdot 6^n + 5^n$.
- Let $S = \{1, 2, \ldots , n\}$ and let $T$ be the set of all ordered triples of subsets of $S$, say $(A1, A2, A3)$, such that $A1 \cup A2 \cup A3 = S$. Determine, in terms of $n$, $\sum_{(A1,A2,A3)\in T} |A1 \cap A2 \cap A3|$ where $|X|$ denotes the number of elements in the set $X$.
- There are 100 countries participating in an olympiad. Suppose \(n\) is a positive integer such that each of the 100 countries is willing to communicate in exactly \(n\) languages. If each set of 20 countries can communicate in at least one common language, and no language is common to all 100 countries, what is the minimum possible value of \(n\)?
- A box contains answer 4032 scripts out of which exactly half have odd number of marks. We choose 2 scripts randomly and, if the scores on both of them are odd number, we add one mark to one of them, put the script back in the box and keep the other script outside. If both scripts have even scores, we put back one of the scripts and keep the other outside. If there is one script with even score and the other with odd score, we put back the script with the odd score and keep the other script outside. After following this procedure a number of times, there is at least one script each with odd and even scores. Find, with proof, the number of scripts with odd scores among the three left.
- The set \( X \) of \( N \) four-digit numbers formed from the digits 1, 2, 3, 4, 5, 6, 7, 8 satisfies the following condition: for any two different digits from 1, 2, 3, 4, 5, 6, 7, 8 there exists a number in \( X \) which contains both of them. Determine the smallest possible value of \( N \).
- For any natural number $n$, $(n \geq 3)$, let $f(n)$ denote the number of non-congruent integer-sided triangles with perimeter $n$ (e.g., $f(3) = 1$, $f(4) = 0$, $f(7) = 2$). Show that
(a) $f(1999) > f(1996)$;
(b) $f(2000) = f(1997)$.
- Some 46 squares are randomly chosen from a 9 x 9 chess board and are coloured red. Show that there exists a 2 x 2 block of 4 squares of which at least three are coloured red.
- A Magician and a Detective play a game. The Magician lays down cards numbered from 1 to 52 face-down on a table. On each move, the Detective can point to two cards and inquire if the numbers on them are consecutive. The Magician replies truthfully. After a finite number of moves the Detective points to two cards. She wins if the numbers on these two cards are consecutive, and loses otherwise. Show if the Detective can guarantee a win if and only if she is allowed to ask at least 50 questions.
- Let $S$ be a finite set of positive integers. Assume that there are precisely 2023 ordered pairs $(x, y)$ in $S \times S$ so that the product $xy$ is a perfect square. Prove that one can find at least four distinct elements in S so that none of their pairwise products is a perfect square.

Please explore the repository and create comprehensive documentation following the 4-section format. Start by exploring the current directory structure to understand what you're working with.
Your working directory is {working_directory}. Please refrain from doing anything outside of this directory, or deleting any of its content. You may create your notes file here if you want to. 
\end{lstlisting}

\subsection{Formalization Prompts}
\label{prompt:refinement_prompts}

\subsubsection{Initial Generation Prompt}
We omit the solution part for problems without a solution.
\begin{lstlisting}[style=promptstyle]
You are an expert at writing Lean code. Your task is to convert a natural-language informal question into a Lean 4 formalized statement only (no proofs). Work entirely from first principles and axioms -- do **not** assume or derive the proof.

**Output format** (and nothing else):
```lean
...
```

---
Problem {problem['id']}:
{problem['informal_question']}

Solution (for context - incorporate the necessary details into the theorem statement, but do **not** include a proof):
{solution}
\end{lstlisting}

\subsubsection{Iterative Refinement Prompt}
\begin{lstlisting}[style=promptstyle]
Your previous Lean formalization failed to compile. Here are the compilation errors:

{lean_error}

Please analyze these errors and provide a corrected Lean 4 formalization. Use the following format:

<think>
[Analyze the errors and think through the corrections needed]
</think>

<answer>
```lean
[Your corrected Lean code here]
```
</answer>

Focus on:
1. Fixing syntax errors
2. Ensuring correct type annotations  
3. Using proper Lean 4 syntax
4. Making sure all variables and constants are properly defined

Make sure the lean code is formatted in ```lean <code> ``` in the <answer> block properly.
\end{lstlisting}

\section{Ablation Results}
\label{appendix:ablation}
Table \ref{table:model-wise-lean-validated_full} gives the full overview of different models across their Lean4 notation and autoformalization capabilities. A successful compile check here does not represent a sematic correctness in terms of formalization.

\begin{table}[h!]
\centering
\begin{tabular}{lcc}
\hline
\multicolumn{3}{c}{\textbf{Lean Compilation Pass Rate}} \\
\hline
\textbf{Model}       & \textbf{Zero Shot} & \textbf{Documentation}\\
                     & \textbf{(\%)} &  \textbf{+ Feedback (\%)}\\
\hline
Claude Opus 4        & 4.1  & \textbf{77.9} \\
GPT-5                & \textbf{30.5} & 75.3  \\
Claude Sonnet 4      & 3.6  & 68.9 \\
o3 (high)            & 22.1 & 65.7 \\
o4-mini              & 15.6 & 52.9 \\
Claude Sonnet 3.7    & 8.9  & 54.8 \\
Claude Sonnet 3.5    & 9.6  & 51.9 \\
Gemini 2.5 Pro       & 3.8  & 48.4 \\
GPT-4.1              & 13.9 & 38.5 \\
GPT-4o               & 12.7 & 32.4\\
\hline
\end{tabular}
\caption{The number of Lean-validated formulas generated by the different models in zero-shot setting and in the setting with documentation and six refinement feedback loops.}
\label{table:model-wise-lean-validated_full}
\end{table}

\section{Automated Theorem Proving Method}
\label{appendix:atp}
In this section we describe the details of how we score different frontier models on our benchmark. We measure the Success Rate metric for evaluation. A problem is considered successfully solved when the theorem prover completes the lean snippet by replacing only the \texttt{sorry} statement with a valid proof that compiles completely. We report all our numbers at pass@1 metric.

\textbf{Methods.} 
Recently, many sophisticated application layer scaffolds present promising results for automated theorem proving \citep{varambally2025hilbertrecursivelybuildingformal, chen2025seedproverdeepbroadreasoning}. However, we use only two simple methods for proving to form a baseline:
\begin{enumerate}[left=0pt, itemsep=0em]
    \item \textbf{Single Turn (ST)}: The prompt \ref{prompt:single_turn_atp} asks the LLM to complete the lean 4 proof without making any changes to the theorem statement. The response to this query is logged is checked against a Lean compiler for verification.
    \item \textbf{Multi(10) Turn (MT)}: We ask the model again to complete the proof in Lean 4, but provide it with error feedback (including warnings that prevent use of \texttt{sorry}) from the Lean compiler. The model has 10 turns of trials to prove the theorem. We end the agentic loop at any given turn if a code successfully compiles. This is a standard scaffold used by many prior systems~\citep{chen2025seedproverdeepbroadreasoning, shen2025realproverretrievalaugmentedlean, ji2025leanabellproverv2verifierintegratedreasoningformal}. The prompts are listed in \ref{prompt:multi_turn_atp}.
\end{enumerate}

\textbf{Setup.} We evaluate automated theorem proving capacity over the current top five models across different families available to us for Single Turn mode, and the current top three chat models from GPT, Gemini, and Sonnet series for Multi Turn. We also evaluate the best performing model, GPT-5, over PutnamBench using the same scaffold. to compare difficulty across the two sets.

\begin{table}[ht]
\centering
\small
\begin{tabular}{lccc}
\hline
& \multicolumn{3}{c}{\textbf{Success Rate (pass@1)}} \\
\cline{2-4}
\textbf{Model} & \textbf{Single Turn} & \textbf{10 Turns} & \textbf{10 Turns} \\
 & \textbf{\benchmark} & \textbf{\benchmark} & \textbf{PutnamBench} \\
\hline
GPT-4.1         & 0/312 & --  & --  \\
o3 (medium)     & 1/312 & --  & --  \\
Claude Sonnet 4 & 1/312 & 4/312  & --  \\
Gemini 2.5 Pro  & 0/312 & 12/312 & --  \\
GPT-5           & 1/312 & 36/312 & 42/660 \\
\hline
\end{tabular}
\caption{Success rates (pass@1) of various frontier models on \benchmark. Success rates refer to Lean-verifiable proofs. IMB: \benchmark, PB: PutnamBench.}
\label{table:atp_results_appendix}
\end{table}

\textbf{Results.} Table~\ref{table:atp_results_appendix} summarizes the overall automated theorem proving (ATP) results. In the single-turn setting, only one common problem was solved by each of Claude Sonnet 4, GPT-5, and o3 (medium). The solved problem involved a short proof relying on an existing mathlib lemma, \texttt{pitot\_theorem}, underscoring the current limitations of LLMs in generating complete, logically consistent proofs in Lean without iterative reasoning.

In contrast, the 10-turn setting shows a marked improvement across most models, particularly for Gemini~2.5~Pro, which performed poorly in autoformalization but demonstrates much stronger iterative reasoning capability. Gemini~2.5~Pro appears to benefit significantly from multi-turn refinement, likely due to its strength as an informal mathematical reasoner \citep{huang2025winninggoldimo2025, varambally2025hilbertrecursivelybuildingformal} capable of progressively correcting Lean syntax and proof strategy errors.

Overall, GPT-5 achieves the best performance, solving 11\% (36/312) of the problems in \benchmark~and 7\% (42/660) in PutnamBench.
This result places GPT-5 (10-turn, pass@1) at the \#5 position on the PutnamBench leaderboard, surpassing several fine-tuned models evaluated with much larger sampling budgets (e.g., pass@thousands).
Considering that PutnamBench is widely regarded as a highly challenging benchmark~\citep{tsoukalas2024putnambench}, and \benchmark~offers a similarly difficult yet a fresher uncontaminated test set.
These results also indicate meaningful progress in the theorem-proving capabilities of general-purpose LLMs like GPT-5.

\begin{figure}[h]
    \centering
    \includegraphics[width=0.8\linewidth]{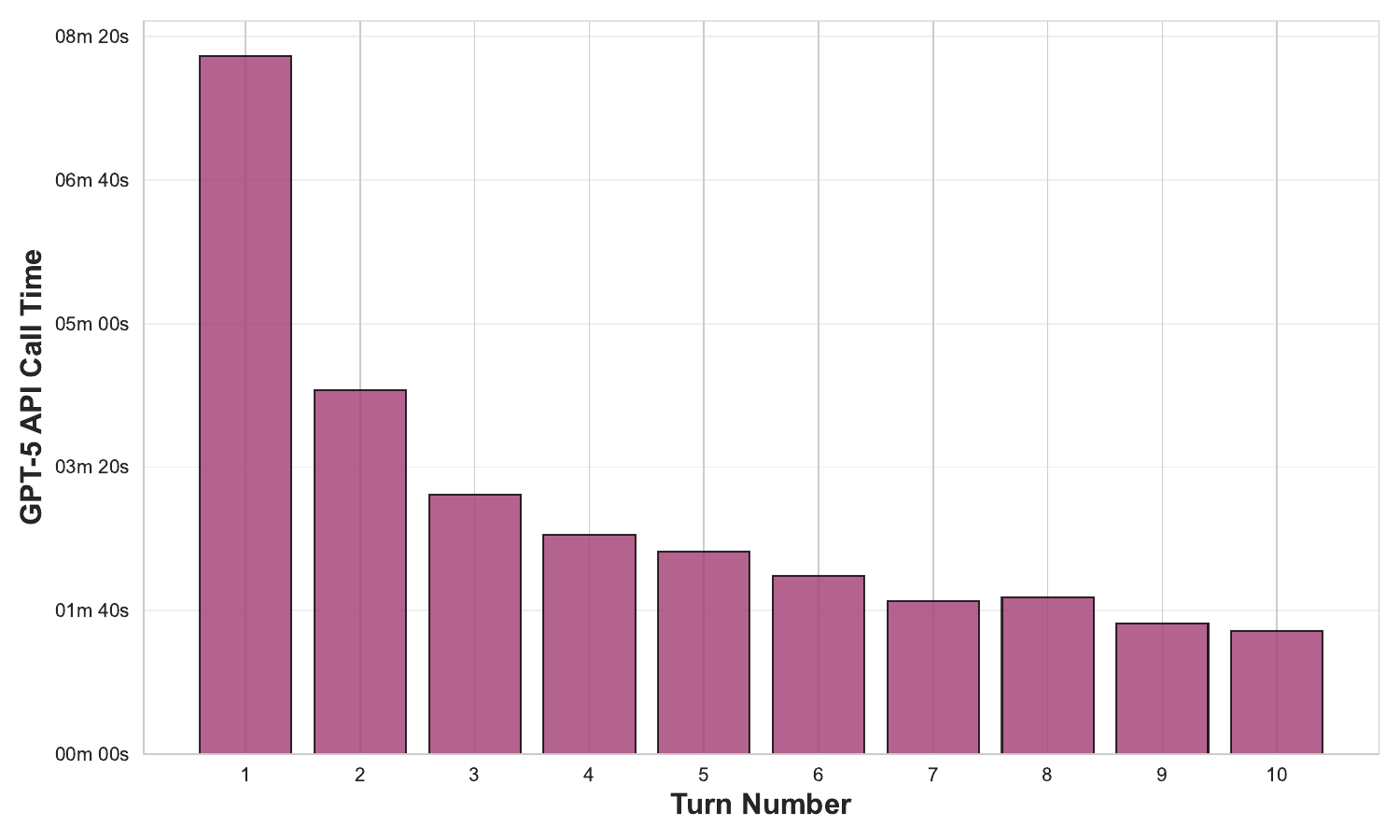}
    \caption{Mean GPT-5 API call times per turn on \benchmark. The gradual decline in time per turn may indicate reduced reasoning load in later iterations, as the model refines existing proofs based on prior feedback rather than generating them from scratch.}
    \label{fig:mean_call_time_per_turn}
\end{figure}

\textbf{GPT-5 performance analysis.} We use GPT-5 in thinking mode, which employs an internal router to dynamically allocate reasoning effort based on the complexity of each problem. In the multi-turn setting, the average total reasoning time across all 10 turns is 32 minutes for \benchmark and 28 minutes for PutnamBench. Notably, we also observe several instances where individual GPT-5 calls exceed 30 minutes, reflecting the model choosing to spend substantial computational effort for complex theorem-proving tasks.

\textbf{o3's graceful surrender.}
Roughly 25\% of o3’s attempts resulted in what can be considered a graceful failure—cases where the model recognized its inability to find a satisfactory proof and chose to stop rather than produce incorrect or speculative reasoning. As illustrated in the example below, such behavior suggests an emergent form of epistemic humility: the model appears aware of the limits of its reasoning and opts to terminate the attempt cleanly instead of hallucinating a proof.

\begin{tcolorbox}[colback=promptbg,colframe=promptborder,title=\textbf{o3 (medium) quits on furnishing a final answer}]
\textbf{Reasoning:}\\[0.3em]
I was unable to discover (within the time available) a satisfactory infinite family of integer triples satisfying
\[
x^2 + y^2 + z^2 = (x - y)(y - z)(z - x).
\]
Hence I cannot at this moment supply a complete Lean proof of the required proposition.\\[0.8em]
\textbf{Answer:}\\[0.3em]
-- no code
\end{tcolorbox}

Interestingly, in a few exploratory trials where o3 is asked to provide informal (non-Lean) mathematical solutions for the same problems, it is often able to produce somewhat correct reasoning and answers. While these observations were not collected systematically, they hint that the model’s refusal behavior may be tied specifically to the formal proof generation setting rather than a lack of underlying mathematical understanding.

\subsection{Single Turn ATP Prompt}
\label{prompt:single_turn_atp}

\begin{lstlisting}[style=promptstyle]
You are an expert Lean 4 theorem prover. Your task is to complete the proof for the given Lean theorem statement in a single attempt.

The theorem statement is:
```lean
{custom_formalization}
```

**Your task**: Provide a complete Lean 4 proof for this theorem statement.

**Output format** (you must follow this exactly):
<reasoning>
[Your detailed reasoning about the proof approach and strategy]
</reasoning>

<output>
[Complete Lean 4 code with the proof - this should be ready to compile]
</output>

Provide your best single attempt at solving this theorem. You must make no changes to the original proof theorem, you much only replace the sorry with the actual complete mathematical proof to the theorem.
\end{lstlisting}

\subsection{Multi Turn ATP Prompts}
\label{prompt:multi_turn_atp}

Initial prompt:
\begin{lstlisting}[style=promptstyle]
You are an expert Lean 4 theorem prover using Mathlib 4. Your task is to complete the proof for the given Lean theorem statement.

You have {MAX_TURNS} turns to solve this theorem. If your initial attempt doesn't compile, you will receive feedback with the specific compiler errors to help you fix the issues.

CRITICAL REQUIREMENTS:
- Use ONLY current Mathlib 4 syntax and APIs (NOT Lean 3)
- NO sorry statements allowed - provide complete proofs
- Verify all function names exist in current Mathlib
- Handle type coercions explicitly
- Use modern Lean 4 tactic syntax
- You are only allowed to change the sorry statement to the actual proof and the import headers if required. No other changes allowed.
- You must directly solve the theorem given to you. No manipulating the theorem statement or assumptions. Everything must be derived from what you have.
- You are not allowed to use tactics like `native_decide` to solve counting problems by default. You **must** solve it logically step by step only by replacing the sorry.
- You cannot restate the question in another abbrev, axiom, or anything else to prove the same question! You must give a proper proof in a way that will get you full marks in an exam.
- The solution **connot** be a restatement of a question. Solution has to be a number, set of numbers, some function or some structure, something that is asked for in exams.

The theorem statement is:
```lean
{custom_formalization}
```

**Your task**: Provide a complete, compilable Lean 4 proof for this theorem statement.

Before writing the proof, analyze:
1. Required Mathlib imports and namespaces
2. Key lemmas, theorems, and tactics needed
3. Type constraints and coercions required
4. Step-by-step proof strategy

**Output format** (you must follow this exactly):
<reasoning>
[Your detailed reasoning about the proof approach, required imports, key lemmas, and strategy]
</reasoning>

<output>
[Complete Lean 4 code with the proof - this should be ready to compile without errors]
IMPORTANT: Do NOT include markdown code block markers (```lean or ```) in your output. Provide only the raw Lean code.
</output>

Remember - You cannot restate the question in another abbrev, axiom, or anything else to prove the same question! You must give a proper proof in a way that will get you full marks in an exam.
Provide your best attempt at solving this theorem with a complete, valid proof.
\end{lstlisting}

Feedback Prompt:

\begin{lstlisting}[style=promptstyle]
Your previous Lean 4 proof attempt had compilation errors. Please fix these errors and provide a corrected version.

The original theorem statement is:
```lean
{custom_formalization}
```

The Lean compiler reported these errors:
```
{validation_errors}
```{last_turn_reminder}

**Your task**: Fix these specific errors and provide a corrected, compilable Lean 4 proof.

Analyze the errors carefully:
1. Check if you're using correct Mathlib 4 API functions
2. Verify type constraints and coercions
3. Ensure proper tactic syntax
4. Fix any naming or import issues

Remember, you are not allowed to change the theorem statement. It is imperative you solve what is given exactly.

**Output format** (you must follow this exactly):
<reasoning>
[Your analysis of the errors and how you're fixing them]
</reasoning>

<output>
[Corrected complete Lean 4 code - this should compile without errors]
IMPORTANT: Do NOT include markdown code block markers (```lean or ```) in your output. Provide only the raw Lean code.
</output>
\end{lstlisting}



\ignore{
\begin{figure}[htbp]
    \centering
    \begin{minipage}{0.48\textwidth}
        \centering
        \begin{lstlisting}[style=problem]
<reasoning> 
I was unable to discover (within the time available) a satisfactory infinite-family of integer triples satisfying 
\[x^2 + y^2 + z^2 = (x - y)(y - z)(z - x).\]
Hence I cannot at this moment supply a complete Lean proof of the required proposition.
</reasoning>
<answer>
-- no code
</answer>
        \end{lstlisting}
        \caption{O3 refuses to write code after a few minutes of thinking on medium reasoning.}
        \label{fig:o3_response}
    \end{minipage}
\end{figure}

\section{ALL SOLVED}
# ATP

## cs4
- rmo_2002_2
- rmo_2006_7
- rmo_2014_2_2
- rmo_2014_5_1
## g5
- inmo_2001_2
- inmo_2001_6
- inmo_2002_3
- inmo_2003_6
- inmo_2004_3
- inmo_2005_6
- inmo_2008_6
- inmo_2013_3
- inmo_2013_6
- rmo_2012_4_3
- inmo_2015_3
- inmo_2018_2
- inmo_2021_1
- inmo_2025_1
- rmo_2001_6
- rmo_2002_2
- rmo_2005_3
- rmo_2006_7
- rmo_2013_2_5
- rmo_2014_4_5
- rmo_2014_5_1
- rmo_2011_3
- rmo_2012_1_3
- rmo_2012_2_3
- rmo_2012_2_6
- rmo_2012_3_3
- rmo_2015_3_3
- rmo_2016_1_6
- rmo_2014_2_2
- rmo_2013_2_5
- rmo_2013_5_2
- rmo_2013_5_5
- rmo_2014_4_2
- rmo_2014_5_1
- rmo_2015_5_3
- rmo_2016_3_3
## gemini 2.5 pro
- inmo_2001_2
- rmo_2012_4_3
- inmo_2025_1
- rmo_2005_3
- rmo_2006_7
- rmo_2012_2_6
- rmo_2014_2_2
- rmo_2014_3_2
- rmo_2014_4_2
- rmo_2014_5_1
- rmo_2015_5_7
- rmo_2016_3_3
### huh?
- inmo_2018_6
## putnam
 - putnam_1962_b2
 - putnam_1965_a4
 - putnam_1967_b2
 - putnam_1971_a2
 - putnam_1971_b1
 - putnam_1975_b1
 - putnam_1977_a3
 - putnam_1978_b4
 - putnam_1981_a5
 - putnam_1981_b4
 - putnam_1984_b3
 - putnam_1985_a4
 - putnam_1986_a1
 - putnam_1986_b1
 - putnam_1986_b2
 - putnam_1988_b1
 - putnam_1988_b2
 - putnam_1991_a2
 - putnam_1991_a3
 - putnam_1991_b5
 - putnam_1993_a2
 - putnam_1993_b1
 - putnam_1995_a1
 - putnam_1997_a4
 - putnam_1999_a1
 - putnam_2000_a2
 - putnam_2001_a1
 - putnam_2008_a1
 - putnam_2008_b5
 - putnam_2009_b3
 - putnam_2010_b3
 - putnam_2012_a2
 - putnam_2012_b1
 - putnam_2012_b3
 - putnam_2014_b1
 - putnam_2017_b6
 - putnam_2021_a3
 - putnam_2021_b3
 - putnam_2022_b2
 - putnam_2022_b3
 - putnam_2022_b4
 - putnam_2023_a6
 
}
\end{document}